\documentclass[lettersize,journal]{IEEEtran}
\usepackage{amsmath,amsfonts}
\usepackage{algorithmic}
\usepackage{algorithm}
\usepackage{array}
\usepackage[caption=false,font=normalsize,labelfont=sf,textfont=sf]{subfig}
\usepackage{textcomp}
\usepackage{stfloats}
\usepackage{url}
\usepackage{verbatim}
\usepackage{graphicx}
\usepackage{cite}
\usepackage{orcidlink} 
\hypersetup{hidelinks} 

\usepackage{amssymb}
\usepackage{multirow}
\begin{document}

\title{Reputation-Driven Asynchronous Federated Learning for Enhanced Trajectory Prediction with Blockchain}

\author{Weiliang Chen$^{\orcidlink{0000-0003-1978-7504}}$, Li Jia$^{\orcidlink{0000-0002-5566-9209}}$, Yang Zhou$^{\orcidlink{0000-0002-2582-9494}}$, Qianqian Ren$^{\orcidlink{0000-0003-1171-7018}}$

\thanks{Manuscript received April 19, 2021; revised August 16, 2021. This work was supported by National Natural Science Foundation of China (62303296, 61773251). \textit{(Corresponding author: Yang Zhou, Qianqian Ren.)}

Weiliang Chen and Qianqian Ren are with the Department of Computer Science and Technology, Heilongjiang University, Harbin 150080, Peoples R China (e-mail: chanweiliang@s.hlju.edu.cn, renqianqian@hlju.edu.cn). 

Li Jia and Yang Zhou are with Shanghai Univ, Sch Mechatron Engn \& Automat, Shanghai 200444, Peoples R China (e-mail: jiali@shu.edu.cn; zhouyang0410@shu.edu.cn). 

Yang Zhou is with Shanghai Univ, Shanghai Artificial Intelligence Lab, Shanghai 201109, Peoples R China (e-mail: zhouyang0410@shu.edu.cn). 
}
}

\markboth{Journal of \LaTeX\ Class Files,~Vol.~14, No.~8, August~2021}%
{Shell \MakeLowercase{\textit{et al.}}: A Sample Article Using IEEEtran.cls for IEEE Journals}


\maketitle

\begin{abstract}
Federated learning combined with blockchain empowers secure data sharing in autonomous driving applications. Nevertheless, with the increasing granularity and complexity of vehicle-generated data, the lack of data quality audits raises concerns about multi-party mistrust in trajectory prediction tasks. In response, this paper proposes an asynchronous federated learning data sharing method based on an interpretable reputation quantization mechanism utilizing graph neural network tools. Data providers share data structures under differential privacy constraints to ensure security while reducing redundant data. We implement deep reinforcement learning to categorize vehicles by reputation level, which optimizes the aggregation efficiency of federated learning. Experimental results demonstrate that the proposed data sharing scheme not only reinforces the security of the trajectory prediction task but also enhances prediction accuracy.
\end{abstract}


\section{Introduction}
\IEEEPARstart{T}{he} rapid advancements in emerging computational and communication technologies within the 5G network have revolutionized the operational paradigms of modern vehicular services and applications, thereby enhancing the overall driving experience\cite{5GAutoTrans}. Autonomous vehicles represent intricate systems that amalgamate various technologies, including perception algorithms\cite{AutoPerceptiona}, path planning\cite{PathPlan}, control theory\cite{VehicleControl}, and vehicle positioning\cite{Autopositioning}. Among the crucial research domains in autonomous vehicles, trajectory prediction\cite{TrajectorySurvey} stands out as it plays an essential role in the decision-making processes of intelligent vehicles. A precise and dependable vehicle trajectory prediction algorithm holds the capability to anticipate potential traffic accidents, thus safeguarding overall vehicular safety by enabling timely preventive measures.

Currently, trajectory prediction in the context of autonomous vehicles is mainly driven by neural network approaches. Recurrent neural networks (RNNs) are leading the way. Researchers have made extensive efforts to develop vehicle trajectory prediction models based on historical trajectory data and environmental information. For instance, GRIP\cite{GRIP} optimizes convolutional layers with graph operations, treating the driving scenario as a graph model. In this way, the interactions between the vehicles are represented as the interrelationships between the nodes of this graph model. The method\cite{Spectral} introduces a novel framework based on spectral clustering, enabling simultaneous prediction of vehicle trajectories and driving behavior. While these studies yield promising results, we must recognize that trajectory prediction faces significant challenges in the real world, especially in traditional telematics environments. In these cases, owners of trajectory data face a vexing problem of data silos. This issue emphasizes the necessity for creative solutions, and federated learning (FL) is a promising method for addressing it.



FL presents a viable strategy for preserving data privacy in distributed settings\cite{FLSvy}. It achieves edge intelligence by imparting knowledge gleaned from decentralized data sources while upholding stringent privacy standards. Moreover, when coupled with the inherent features of blockchain technology\cite{FLblockchain}, such as tamper resistance, anonymity, and traceability, it establishes a robust foundation for efficient and reliable data sharing among participants. Even in situations where trust among participants is limited, users are required to transmit model parameters to a central server for aggregation. This not only facilitates parallel learning to enhance the global model but also ensures data integrity, maintains anonymity, and provides traceability. Nonetheless, as autonomous driving continues to advance rapidly and the data generated by vehicles becomes increasingly complex, the task of trajectory prediction still confronts several pressing challenges that require resolution:
\begin{enumerate}
\item \textbf{Lack of complete trust in multiple parties.} 
Many current data sharing schemes with central administrators pose an increased risk of data leakage. Administrators must manage large volumes of aggregated data from multiple parties, including unknown or raw sources.
\item \textbf{Uncertainty in participant data quality.}
Due to vehicle mobility and unreliable communications, the data sharing environment is highly fluid, with new unverified data being created constantly. When data providers share malicious or redundant data, it can significantly bias the entire prediction process.

\item \textbf{Extensive prediction delays.}
In the context of updating dynamic vehicle data, addressing aggregation challenges caused by delays in heterogeneous vehicle data is crucial. The conventional approach, which relies solely on federated learning-enabled blockchain technology, requires coordination among Roadside Units (RSUs) at each timestamp before uniformly uploading data for global aggregation.
\end{enumerate}

To systemically address these limitations, we introduce a novel trajectory prediction approach RAFLTP based on graph neural networks to address the aforementioned challenges. We design an asynchronous FL strategy based on reputation evaluation to quantify vehicle reputation through interactions between vehicles. Furthermore, data providers release data structures under the constraints of differential privacy. Due to the excellent performance of deep reinforcement learning (DRL) in dealing with dynamic stochastic decision problems, many studies use DRL to improve the performance and efficiency of FL. For instance, deep Q-learning network (DQN) and FL are applied jointly to address challenges in edge computing, including task offloading \cite{DQNFL}, caching, and communication issues \cite{DQNFLcach}. Additionally, deep deterministic policy gradient (DDPG) is employed to select device nodes with high data quality for improved model aggregation and reduced communication costs \cite{DDPGIIOT,DDPGAuto}. Moreover, proximal policy optimization (PPO) is utilized to determine the optimal task scheduling strategy \cite{PPORobot}. It is worth noting that both DQN and PPO are more adept at solving discrete action space problems. Due to its excellent stability and efficiency \cite{schulman2017proximal}, PPO is our primary choice for achieving the vehicle grouping task to minimize the FL cost.
The contributions of this paper can be summarized as follows.
\begin{itemize}
\item{We present an interpretable reputation-based framework for secure data sharing in trajectory prediction tasks. We design a reputation reward mechanism that replaces the traditional loss function with reputation values computed from vehicle trajectory data, driving the method components to cooperate and promoting federated learning to train more instructive global models.}
\item{We propose a reputation-driven asynchronous FL scheme adapted to dynamic and heterogeneous vehicle networks. This scheme is complemented by a reputation-enhanced PPO algorithm that groups vehicle trajectory models based on reputation values to reduce the FL cost. High-reputation vehicle clusters, characterized by high data quality and similar trajectory graph models, are prioritized for deep global federated learning aggregation.}
\item{We introduce differential privacy techniques to enhance the privacy and security of shared trajectory graph models and vehicle reputation values while preserving the graph structure and interactions among vehicles.}
\item{Extensive experiments are conducted on two large-scale vehicle trajectory datasets, namely NGSIM and ApolloScape, with different real-world scenarios, comparing our approach with other related models. The results demonstrate that our data sharing scheme effectively ensures data security, enhances trajectory prediction accuracy, and adapts well to various traffic scenarios. Moreover, comprehensive ablation studies confirm the effectiveness of our method components.}
\end{itemize}

The rest of the paper is organized as follows. Section II presents related work. Section III gives the problem definition. Sections IV and V elaborate on the proposed model. Section VI presents an extensive experiment evaluation. Section VII concludes the paper.

\section{RELATED WORK}

Trajectory prediction spans multiple domains, incorporating statistics\cite{RWstatistics}, signal processing\cite{RWsignalPro}, and control systems engineering\cite{VehicleControl}. Contemporary trajectory prediction techniques often rely on data-driven approaches, particularly deep learning methods. CS-LSTM\cite{CS-LSTM} and Social-STGCNN\cite{Social-STGCNN} combine neural networks and graph-based models to handle vehicle interactions. Some deep learning methods\cite{DAGNet,trafficpredict,RWchoupredicting} have been employed for trajectory prediction. However, they primarily focus on local road interactions. In contrast, graph-based methods such as GRIP\cite{GRIP} and certain traffic density prediction techniques are capable of accommodating interactions without geographical constraints, ensuring a comprehensive view. Spectral\cite{Spectral} simulates the entire world, allowing graph models to capture diverse temporal correlations. Real-world traffic scenarios often face data silos, contrasting with the Internet of Vehicles (IoV)\cite{IoV1}, an emerging paradigm that promotes high-quality services by enabling vehicles to share road-related information within in-vehicle networks. Data sharing\cite{RWdataQA} is integral to enhancing driving experiences and Internet of Thing (IoT) services, with data quality hinging on vehicle reputation\cite{VTruMA}, considering the potential for incorrect or irrelevant information stemming from defective sensors, corrupted firmware, or selfish motives. Therefore, developing a mechanism for quantifying vehicle reputation based on inter-vehicle interactions is pivotal\cite{RWCloudTru}.

Trust management systems in vehicular networks enable vehicles to assess information trustworthiness and provide network operators with a basis for incentives or penalties\cite{rewardveh}. Centralized trust systems often struggle to meet stringent quality of service (QoS) requirements due to short decision times and central server reliance\cite{centralisedTru}. Distributed trust systems, where vehicles or roadside units manage trust locally, reduce network infrastructure interactions but face challenges due to variable vehicle capabilities and network dynamics\cite{DistributedRep}. Some studies involve roadside units in trust management\cite{VANETs}, but RSUs can be half-trusted, leading to inconsistent services\cite{RSUsecurity}. Various schemes exist, including trust calculation upon data reception and joint privacy-reputation consideration\cite{RayaTru}. These approaches require vehicles to manage trust values, which may be inaccurate due to limited observations or failures\cite{LiChiganTru}. Blockchain technology, known for its security features, has gained attention for vehicular data sharing\cite{BCVN2}. Combining AI and blockchain enhances resource sharing under 5G conditions. Despite efforts like the Bayesian inference model on a public blockchain\cite{VTruMA}, creating a decentralized, reliable, and consistent vehicular trust management system remains challenging.

FL\cite{FLSTR} offers privacy-preserving edge intelligence in distributed scenarios, allowing users to maintain their data while sharing model parameters with a server for collaborative global model learning. Traditional synchronous FL results in high communication costs and idle times\cite{FLURC,HanTPFL}. To improve efficiency, an asynchronous mini-batch algorithm was proposed, addressing optimization problems with multiple processors\cite{asynchronousFL}. However, existing FL methods like Stochastic Gradient Descent (SGD) or Deep Neural Networks (DNN) may risk exposing sensitive data. Most data sharing schemes involve large amounts of aggregated data, including fresh, malicious, or redundant data\cite{FL_TIFS}. Differential privacy\cite{DP}, unlike prior models such as k-anonymity\cite{Kanonymity} and l-diversity\cite{Ldiversity}, offers reliable privacy guarantees, defending against various privacy attacks. A machine learning differential privacy\cite{DPSry} has been proposed that releases data structures under the constraints of differential privacy rather than directly disclosing queries and answers.

\section{SYSTEM MODEL}
In this paper, based on the inefficiency and poor security of current trajectory prediction methods, we propose a new asynchronous FL model for trajectory prediction, which effectively addresses the problems of inefficient training and privacy leakage in existing methods.
\subsection{Formulation of Trajectory Prediction}
\begin{figure}[!t]
\centering
\includegraphics[width=3.4in]{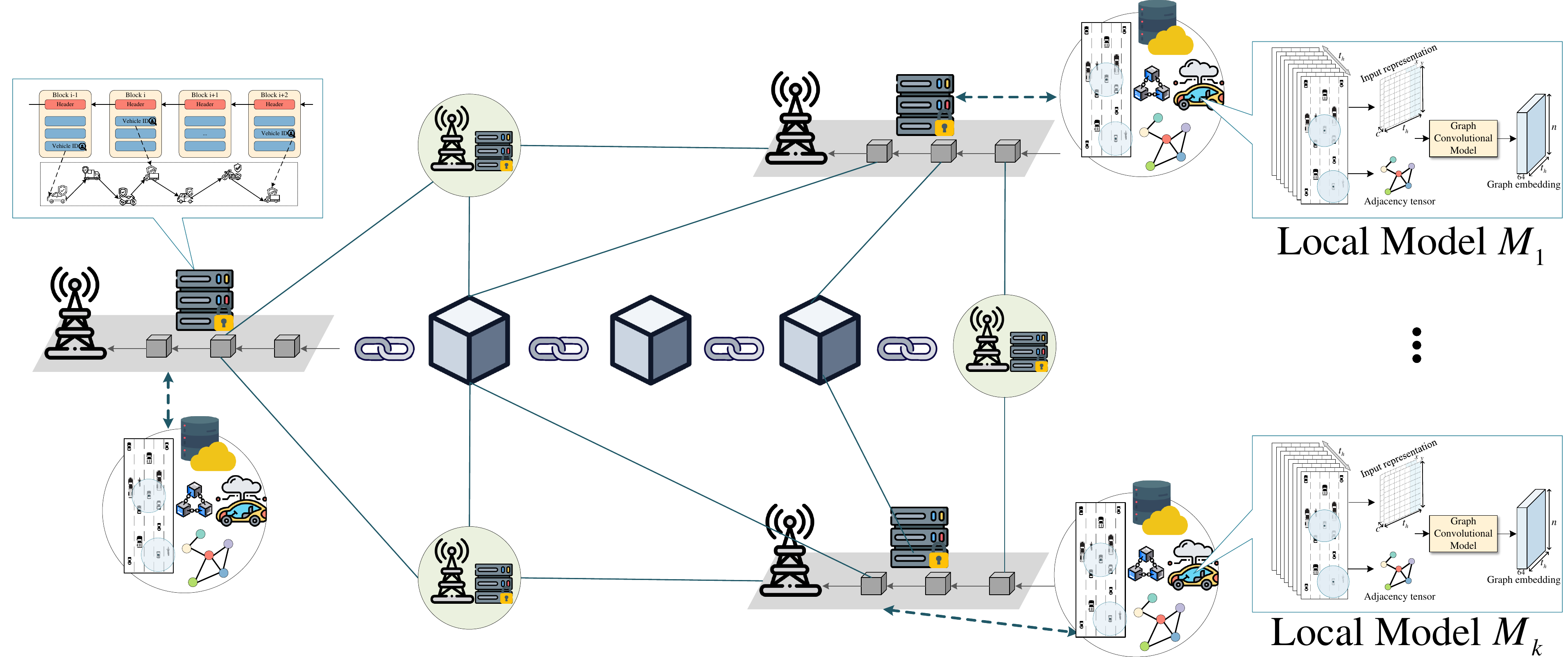}
\caption{Architecture of secure data sharing solution.}
\label{datasharing}
\end{figure}

To begin with, we propose integrating the task of trajectory prediction into a practical scenario of distributed data sharing that involves multiple participants. Each participant has their own data and is willing to share it to realize the collaborative task, but we cannot exclude the existence of malicious sharers who try to interfere with the trajectory prediction task. In addition, the vehicular network comprises vehicles, RSUs, and Macro Base Stations (MBSs), primarily utilizing vehicle-to-vehicle (V2V) and vehicle-to-RSU (V2R) communication. RSUs are equipped with Mobile Edge Computing (MEC) servers, providing computational and storage capabilities. The proposed data sharing architecture comprises a permissioned blockchain module and a FL module, illustrated in Fig.~\ref{datasharing}. The permissioned blockchain handles retrieval and data-sharing transactions. We utilize the permissioned blockchain solely for retrieving relevant data and managing data accessibility, without recording the raw data.

We consider $N$ vehicle participants and a joint dataset $D = \{D_1, D_2, \ldots , D_n\}$. For any vehicle $v_{\text{req}} \in v_i$ that participates in the request for data sharing, its local dataset $D_i \in D$, which contains the historical trajectory $X\in \mathbb{R}^{\tau \times n \times 2}$ of all observations at time step observations at time step $\mathcal{T}$:
\begin{equation}
X=\left[\mathcal{P}^{(1)}, \mathcal{P}^{(2)}, \cdots, \mathcal{P}^{\left(\mathcal{T}\right)}\right]\label{eq:P^t}
\end{equation}
where $\mathcal{P}^{(t)}=\left[\left[x_0^t, y_0^t\right],\left[x_1^t, y_1^t\right], \ldots,\left[x_n^t, y_n^t\right]\right] \in \mathbb{R}^{n \times 2}$ are the coordinates of all observations at time $t$ and $n$ is the number of observations. We employ a coordinate system based on the measured relative values of the self-vehicle with reference to \cite{grip2}. And the output $\mathcal{Y}^{\prime} \in \mathbb{R}^{T \times n \times 2}$, which predicts the future positions of all observations from time step $\mathcal{T} + 1$ to $\mathcal{T} + \mathcal{T}_f$:
\begin{equation}
\mathcal{Y}^{\prime}=\left[\mathcal{P}^{\left(\mathcal{T}+1\right)}, \mathcal{P}^{\left(\mathcal{T}+2\right)}, \cdots, \mathcal{P}^{\left(\mathcal{T} + \mathcal{T}_f\right)}\right]
\end{equation}
where $\mathcal{T}_f$ is the predicted range.

\subsection{Our Proposed Architecture And Workflow}
\begin{figure*}[!t]
\centering
\setlength{\abovecaptionskip}{-0.2cm}
\includegraphics[width=6.8in]{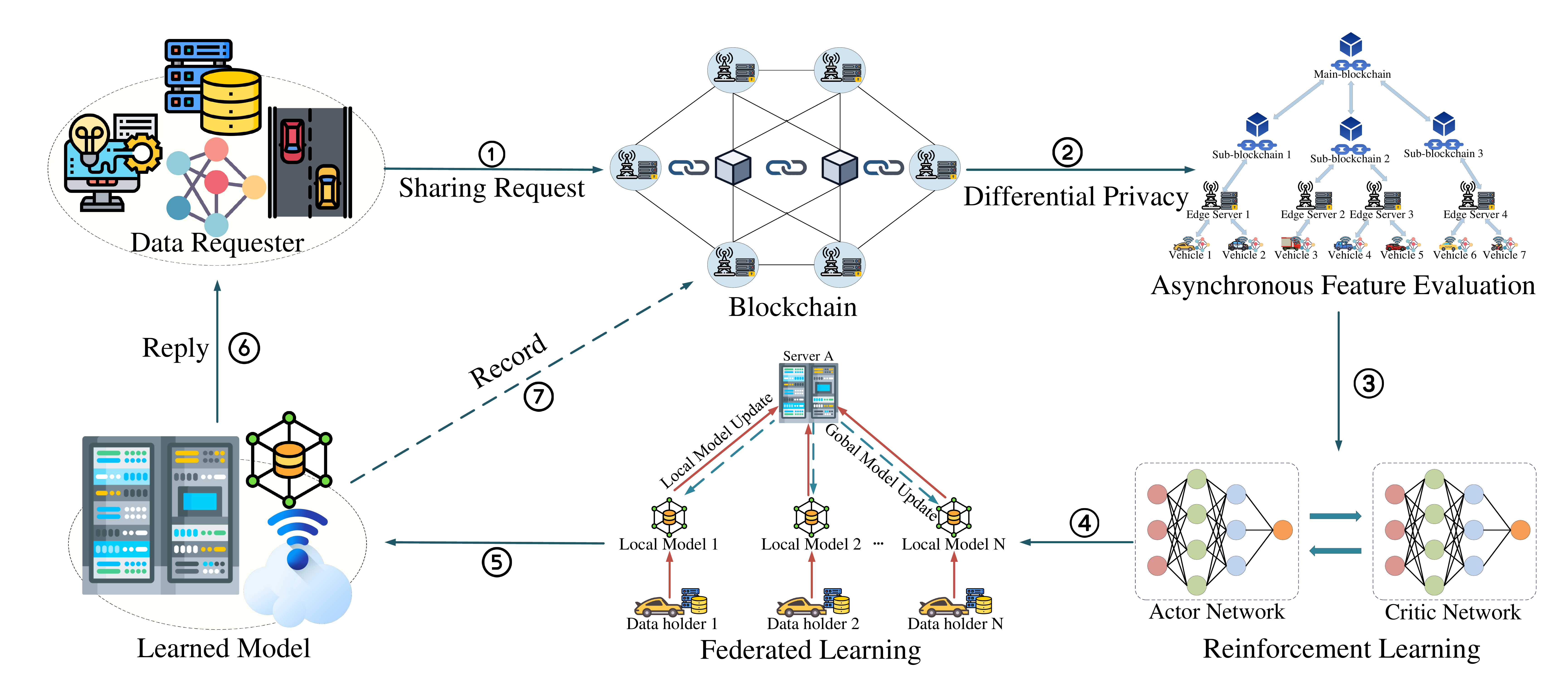}
\caption{Working mechanisms for the proposed methodology.}
\label{ARI}
\vspace{-0.2cm}
\end{figure*}

To improve computational and storage efficiency while facing limited resources, and to ensure secure and robust vehicle interactions, we implement a graph-based approach for model training and data sharing. Initially, vehicles must undergo authentication to become legitimate nodes in the permissioned blockchain, obtaining certificates for participating in V2V data sharing, and enabling them to download the global model locally for training purposes. Subsequently, vehicles transmit data sharing requests to nearby super nodes (such as MBS and RSU) for processing. Upon verifying their public keys, the super nodes commence the collection of weighted graph models perturbed with differential privacy noise from vehicles within the community for federated learning training. Participating vehicles compute reputation values based on the original data to assess the quality of their trajectory data. These values are recorded as shared transactions and in the local models of other participating vehicles within each vehicle's local Directed Acyclic Graph (DAG). To mitigate the global aggregation latency and optimize the aggregation results, we use the PPO algorithm to group vehicles based on the reputation evaluation mechanism. Therefore, the high-quality vehicle clusters receive priority in deep aggregation. The super nodes collaborate to train the global model $M$ through asynchronous FL and broadcast the results to all committee nodes, which are charged with driving the consensus process for the permissioned blockchain. The committee nodes collect the transactions into blocks and verify the blocks to further add them to the permissioned blockchain. The vehicles optimize their local models by integrating the global model, thereby achieving increased accuracy in predicting trajectories. Fig.~\ref{ARI} shows the working mechanism of our scheme.

\section{Reputation-Based Hybrid Blockchain For Data Sharing}
Considering resource constraints and privacy concerns of vehicle users, we propose a data sharing scheme, which involves the collaborative learning of a federated graph neural network model across decentralized parties to share well-trained model parameters instead of raw data.

\subsection{Graph-Based Trajectory Modeling}
Our approach transforms object trajectories into weighted graphs for interaction modeling. Referring to the trajectory prediction model GRIP++\cite{grip2}, each object trajectory is organized as a 3D array, computed for improved speed to enhance prediction accuracy. We use undirected graphs $G=\{V, E\}$ to depict the interactions between objects. The input data is processed by graph convolution network (GCN). Graph operations handle spatial interactions and temporal convolution captures time-based features. Both the encoder and decoder used for the trajectory prediction model are two-layer GRU (Gated Recurrent Unit) networks. Based on this, $G=\{V, E\}$ is converted into a weighted graph of vehicle trajectories. The set of nodes $V$ is defined as $V=\left\{v_{i t} \mid i=1, \cdots, n, t=1, \cdots, \mathcal{T}\right\}$, with $n$ denoting the total number of objects observed in the scene. Each node $v_{i t}$ contains its own weights $w_i^n$. The weights $w_{v_{i t}}$ are determined based on the similarity of trajectories between vehicles, including direction, speed, and tilt angle, as described in the following subsection. Subsequently, $G$ is serialized into ordered vectors that are then mapped onto linear vectors. These maps are combined to form a global vehicle network graph $\mathcal{G}=\left\{G_1 \cup G_2, \dots, \cup G_n\right\}$. For the overall graph $\mathcal{G}=\{\mathcal{V}, \mathcal{E}\}$, the number of representative vertices is denoted as $k$. Afterwards, the size of the normalization property of the vertices will be $k$ and the size of the normalization property of the edges will be $k \times(k-1) / 2$. Eventually, the normalized vector $Seq=\mathcal{V} \cup \mathcal{E}=\left\{\mathcal{V}_1, \ldots, \mathcal{V}_k\right\} \cup$$\left\{\mathcal{E}_1, \mathcal{E}_2, \ldots, \mathcal{E}_{k(k-1) / 2}\right\}$.

After that, we use cosine similarity as a distance function to compare vehicle matrices sequentially in temporal order, and confidence levels above 80\% will fail validation, as well as new weighted graphs added in future moments with confidence levels above a preset threshold will fail validation. By using weighted graphs and the PPO algorithm, the set of vehicles $\left\{V_1, \ldots, V_n\right\}$ is divided into high and low reputation groups. Through this differential privacy federation learning mechanism, instead of sharing local vehicle model parameters, the shared graph neural network model is used to retrieve and compare confidence levels thus filtering out the series of graphs with high trustworthiness as well as low similarity, which prevents the model parameters from leaking the vehicle trajectory information and improves the efficiency of the next global aggregation.

\subsection{Differential Privacy-Enhanced Data Sharing}
Our objective is to develop a secure mechanism for data sharing in trajectory prediction scenarios that intelligently facilitates data sharing among distributed multi-users while effectively safeguarding data privacy. Our approach involves considering $\mathcal{N}$ parties or data holders, along with a joint data set $\mathbb{D}$. For each party $P_i$, there exists a local data set $\mathbb{D}_i$ in $\mathbb{D}$. All $\mathcal{N}$ parties unanimously agree to share data without compromising confidential information. Let $R=\left\{r_1, r_2, \ldots, r_m\right\}$ represent requests for data sharing. Requesters submit queries $r_i$, and we provide computed results rather than raw data to fulfill sharing requirements. Finally, the trained global model $M$ is returned to the committee node. Recipients can leverage the received global aggregation model to respond to data sharing requests locally.

Considering the large size and sensitivity of the data, inspired by previous work\cite{Kademlia}, we utilize the blockchain for data retrieval, while the actual data remains stored on local vehicles. We combine graph neural networks and differential privacy for quality verification. Using a graph representation of the raw data improves computational and storage efficiency under resource constraints, preserving more structural and contextual information for validation while also preventing privacy leakage. Once new data providers join, their uniqueness (vehicle ID) is recorded on the blockchain, alongside an overview of their data, including trajectory data, vehicle types, and data sizes. All data profiles from multiple participants are recorded as transactions and verified by blockchain nodes using Merkle trees\cite{Merkle}. Each data sharing event is also stored as a transaction on the blockchain. All participants (denoted as $\left\{P_1, P_2, \ldots, P_n\right\}$) are selected through multi-party searches in the blockchain and are divided into communities based on their reputation values. These communities consist of members with similar data quality. Given the limited communication resources of IoT devices, the retrieval process should also consider the matrix cosine similarity between the two participants. When a user submits a request to share data with a nearby node $P_i$, all nodes within the identical community as $P_i$ broadcast this request to other vehicles that are observing the target during that timestamp to commence the retrieval process. The process is recursively executed until all trajectory graphs for that timestamp have been executed. At the end of the retrieval process, we will procure a collection of vehicles $P_s \subseteq \mathbb{P}$ that pertain to the request. These vehicles exhibit structural stability, the absence of anomalous data, and minimized redundancy in trajectory graphs within the community, addressing the constraints posed by limited communication resources.
To ensure privacy protection for the shared model, we employ differential privacy for each local model $m_i$ using the Laplace mechanism. Here, we add calibrated noise with sensitivity $s$ to the local data to train $\hat{m}_i$, expressed as:
\begin{equation}
\hat{m}_i = m_i + Laplace(s / \epsilon),
\end{equation}
where $\epsilon$ denotes the privacy budget. Subsequently, participant $P_i$ transmits the model $\hat{m}_i$ as a blockchain transaction broadcast to other participants for federated learning. Upon receiving $\hat{m}_i$, participant $P_{i+1}$ trains a new local data model $\hat{m}_{i+1}$ based on the received $\hat{m}_i$ and its local data, and subsequently broadcasts $\hat{m}_{i+1}$ to other participants. This iterative process continues among participants. Finally, a global model $M$ is generated, given by $\mathcal{M} = \left\{\hat{m}_1 \cup \hat{m}_2 \ldots \cup \hat{m}_n\right\}$.
\begin{figure}[!t]
\centering
\includegraphics[width=2.5in]{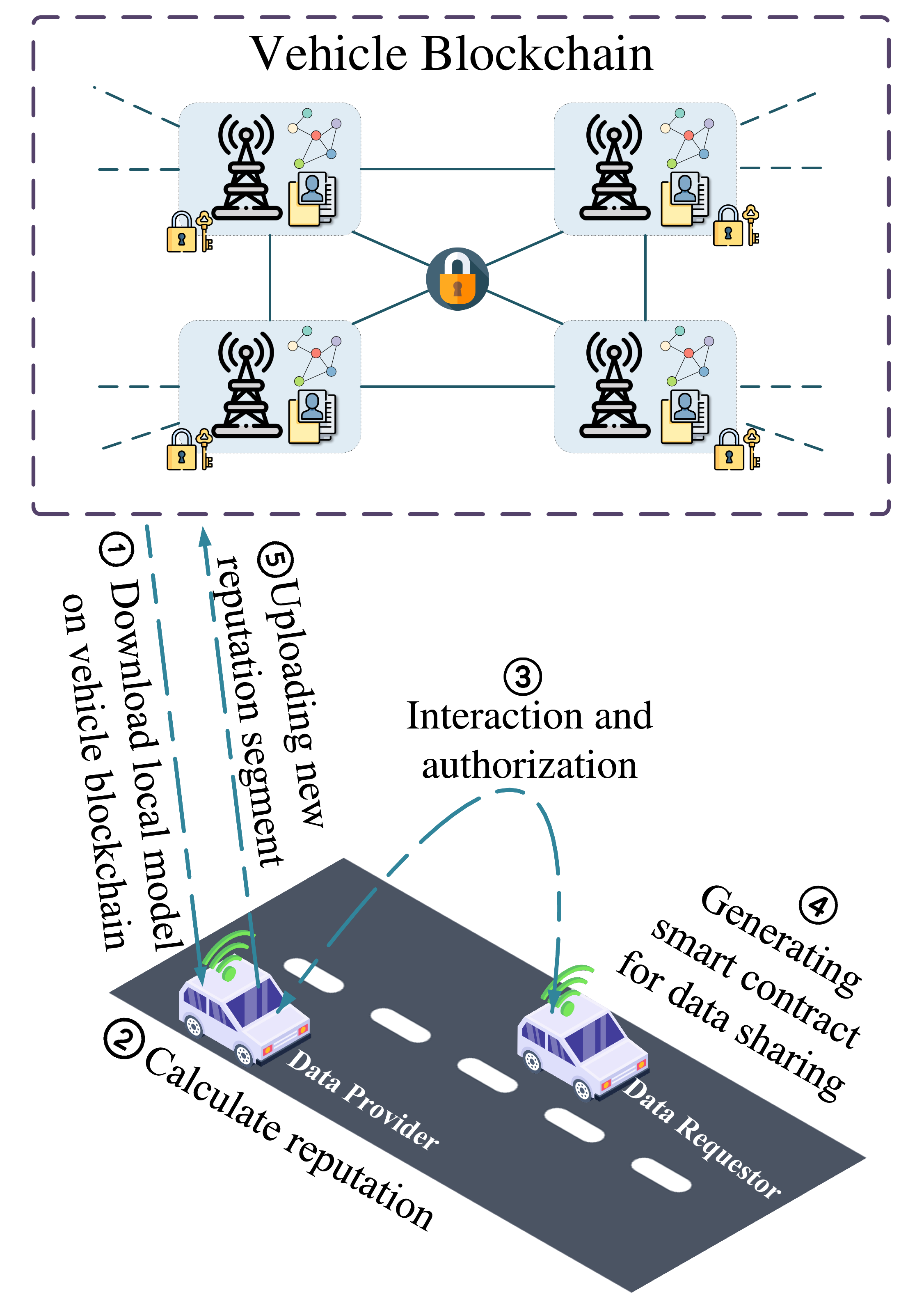}
\caption{Reputation-Aware Data Sharing Mechanism}
\label{reputation}
\end{figure}

\subsection{Trajectory Similarity-Based Reputation-Aware Data Sharing}
The use of Mean Absolute Error (MAE) as an evaluation metric may not be practical in real-world trajectory prediction scenarios. We employ a novel evaluation metric: the reputation value derived from trajectory similarity among vehicles. The locally collected vehicle data is spatially correlated and exhibits a spatial extent. Incorporating trajectory similarity into reputation calculations when sharing data between vehicles enables location awareness and enhances data relevance. The more similar the trajectories, the more relevant the shared data from the data provider, resulting in improved quality, accuracy, and reliability of the shared data. Trajectory coefficients of a vehicle are represented as $v=\{$velocity, position, orientation$\}$. The weights of each coefficient in $v$ are $\rho_1$, $\rho_2$, and $\rho_3$, and $\rho_1+\rho_2+\rho_3=1$. The similarity between two trajectory segments, $\mathcal{J}_i$ and $\mathcal{J}_j$, of vehicles $i$ and $j$, can be expressed as $\mathcal{SIM}\left(\mathcal{J}_i, \mathcal{J}_j\right)$. This is calculated using the Eq.(\ref{Eqsim}),
\begin{equation}
\label{Eqsim}
\mathcal{SIM}\left(\mathcal{J}_i, \mathcal{J}_j\right)=1-\mathcal{DISS}\left(\mathcal{J}_i, \mathcal{J}_j\right)
\end{equation}
The normalized dissimilarity for $L_i$ and $L_j$, denoted as $\mathcal{DISS}\left(\mathcal{J}_i, \mathcal{J}_j\right)$, is used to calculate this similarity, as Eq.(\ref{Eqdiss}),
\begin{equation}
\label{Eqdiss}
\begin{aligned}
\mathcal{DISS}\left(\mathcal{J}_i, \mathcal{J}_j\right)= & \rho_1 \operatorname{velocity}\left(\mathcal{J}_i, \mathcal{J}_j\right)+\rho_2 \operatorname{position}\left(\mathcal{J}_i, \mathcal{J}_j\right) \\ &
+\rho_3 \operatorname{orientation}\left(\mathcal{J}_i, \mathcal{J}_j\right).
\end{aligned}
\end{equation}
We assert that $\mathcal{DISS}\left(\mathcal{J}_i, \mathcal{J}_j\right)$ relies on the disparity in velocity, position, and orientation of the two trajectory segments. The velocity difference between the two segments of the trajectory is given by
\begin{equation}
\operatorname{velocity}\left(\mathcal{J}_i, \mathcal{J}_j\right)=\frac{\left|W_{\mathrm{ave}}\left(\mathcal{J}_i\right)-W_{\mathrm{ave}}\left(\mathcal{J}_j\right)\right|}{\max \left[W\left(\mathcal{J}_i\right), W\left(\mathcal{J}_j\right)\right]}
\end{equation}
where $W\left(\mathcal{J}_i\right)$ and $W\left(\mathcal{J}_j\right)$ are the velocities of vehicles $i$ and $j$ during their trajectory segments, respectively. $W_{\text {ave }}\left(\mathcal{J}_i\right)$ and $W_{\text {ave }}\left(\mathcal{J}_j\right)$ are the average velocities of these two vehicles. The position$\left(\mathcal{J}_i, \mathcal{J}_j\right)$ shows the variance in the position between the trajectory segments. The number of sampled points for $\mathcal{J}_i$ and $\mathcal{J}_j$ during the time window $T$ are denoted as $e$ and $k$, respectively. The set of sampled points in temporal order are $\left\{P_{i 1}, P_{i 2}, \ldots, P_{i e}\right\}$ and $\left\{P_{j 1}, P_{j 2}, \ldots, P_{j k}\right\}$. We employ the Longest Common Sequence (LCS) method to measure the similarity of trajectory segments. For trajectory segments $\mathcal{J}_i$ and $\mathcal{J}_j$, the LCS is described as $\operatorname{LCS}\left(\mathcal{J}_i, \mathcal{J}_j\right)=\left\{P_{i e}=\right.$$\left.P_{j k} \mid e=k\right\}$, where $e \in\{1,2, \ldots, E\}, k \in\{1,2, \ldots, K\}$. Therefore, the equation for the difference in position between the trajectory segment position$\left(\mathcal{J}_i, \mathcal{J}_j\right)$ is provided as follows: 
\begin{equation}
\operatorname{position}\left(\mathcal{J}_i, \mathcal{J}_j\right)=\frac{\max (e, k)-\operatorname{num}\left[\operatorname{LCS}\left(\mathcal{J}_i, \mathcal{J}_j\right)\right]}{\max (e, k)}
\end{equation}
where $\operatorname{num}\left[\operatorname{LCS}\left(\mathcal{J}_i, \mathcal{J}_j\right)\right]$ is the number of points in the LCS of the trajectory segments $\mathcal{J}_i$ and $\mathcal{J}_j$. The directory difference between the two trajectory segments is the angle between the two trajectory segments. Here, we employ $\varphi$ as the angle between the trajectories $\mathcal{J}_i$ and $\mathcal{J}_j$ to represent the direction.  More precisely, 
\begin{equation}
\operatorname{orientation}\left(\mathcal{J}_i, \mathcal{J}_j\right)=\left\{\begin{array}{l}
\frac{\sin \varphi}{2}, 0<\varphi \leq \frac{\pi}{2} \\
\frac{1}{2}+\frac{\left|\sin \left(\varphi+\frac{\pi}{2}\right)\right|}{2}, \frac{\pi}{2}<\varphi \leq \pi
\end{array}\right.
\end{equation}

\subsubsection*{1) Local Opinions for Subjective Logic}
In the case of two vehicles, the reputation between $v_i$ and $v_j$ can be formally described as a local opinion vector $\omega_{i \rightarrow j}:=\left\{r_{i \rightarrow j}, d_{i \rightarrow j}, u_{i \rightarrow j}\right\}$, where $r_{i\rightarrow j}$, $d_{i\rightarrow j}$, and $u_{i\rightarrow j}$ denote trust, distrust, and uncertainty, respectively. $r_{i\rightarrow j}+d_{i\rightarrow j}+u_{i\rightarrow j}=1$, and $r_{i\rightarrow j}$,$d_{i\rightarrow j}$,$u_{i\rightarrow j}\in[0,1]$. Here, 
\begin{equation}
\left\{\begin{array}{l}
r _{i \rightarrow j}=\left(1-u_{i \rightarrow j}\right) \frac{\alpha}{\alpha+\beta} \\
d_{i \rightarrow j}=\left(1-u_{i \rightarrow j}\right) \frac{\beta}{\alpha+\beta} \\
u_{i \rightarrow j}=1-s_{i \rightarrow j}
\end{array}\right.
\end{equation}
where $\beta$ represents the number of negative events, and $\alpha$ represents the number of positive events. The uncertainty of the local opinion vector $u_{i\rightarrow j}$ depends on the quality of communication between vehicles $i$ and $j$, and the communication quality $s_{i\rightarrow j}$ is the probability of successful transmission of packets of data sharing requests during communication. According to $\omega_{i\rightarrow j}$, the reputation value $\mathcal{R}_{i\rightarrow j}$ denotes the expected belief of $v_i$ in providing true relevant data to $v_j$, which can be expressed as
\begin{equation}
\mathcal{R}_{i \rightarrow j}=r_{i \rightarrow j}+\gamma u_{i \rightarrow j}
\end{equation}
The constant $\gamma$ is assigned by the vehicle and measures how much uncertainty affects the reputation of the vehicle. It is recommended to initially set $\gamma$ to 0.5.

\subsubsection*{2) Combine Local Opinions With Suggested Opinions}
Subjective opinions of neighboring vehicles should be taken into account when considering the requesting vehicle. We combine the subjective opinions of different recommenders into a single opinion, named according to the weight of each opinion. These opinions are combined to form the following common opinion  $\omega_{x \rightarrow j}^{\mathrm{rec}}:=$$\left\{r_{x \rightarrow j}^{\mathrm{rec}}, d_{x \rightarrow j}^{\mathrm{rec}}, u_{x \rightarrow j}^{\mathrm{rec}}\right\}$, as follows,
\begin{equation}
\left\{\begin{array}{l}
r_{x \rightarrow j}^{\mathrm{rec}}=\frac{1}{\sum_{x \in X} \delta_{x \rightarrow j}} \sum_{x \in X} \delta_{x \rightarrow j} b_{x \rightarrow j} \\
d_{x \rightarrow j}^{\mathrm{rec}}=\frac{1}{\sum_{x \in X} \delta_{x \rightarrow j}} \sum_{x \in X} \delta_{x \rightarrow j} d_{x \rightarrow j} \\
u_{x \rightarrow j}^{\mathrm{rec}}=\frac{1}{\sum_{x \in X} \delta_{x \rightarrow j}} \sum_{x \in X} \delta_{x \rightarrow j} u_{x \rightarrow j}
\end{array}\right.
\end{equation}
where $x \in X$ represents the set of neighboring vehicles that interact with $v_j$. The data requestor has a local opinion after receiving the shared data from the data provider. In order to avoid spoofing, this local opinion should still be taken into account when forming the final opinion. The final opinion formation for $\omega_{x \rightarrow j}^{\text {final }}:=\left\{r_{x \rightarrow j}^{\text {final }}, d_{x \rightarrow j}^{\text {final }}, u_{x \rightarrow j}^{\text {final }}\right\}$, where $r_{x\rightarrow j}^{\mathrm{final\ }}$, $d_{x\rightarrow j}^{\mathrm{final\ }}$ and $u_{x\rightarrow j}^{\mathrm{final\ }}$ are calculated as
\begin{equation}
\left\{\begin{array}{l}
r_{i \rightarrow j}^{\text {final }}=\frac{r_{i \rightarrow j} u_{x \rightarrow j}^{\text {rec }}+r_{x \rightarrow j}^{\text {rec }} u_{i \rightarrow j}}{u_{i \rightarrow j}+u_{x \rightarrow j}^{\text {rec }}-u_{x \rightarrow j}^{\text {rec }} u_{i \rightarrow j}} \\
d_{i \rightarrow j}^{\text {final }}=\frac{d_{i \rightarrow j} u_{x \rightarrow j}^{\text {rec }}+d_{x \rightarrow j}^{\text {rec }} u_{i \rightarrow j}}{u_{i \rightarrow j}+u_{x \rightarrow j}^{\mathrm{rec}}-u_{x \rightarrow j}^{\text {rec }} u_{i \rightarrow j}} \\
u_{i \rightarrow j}^{\text {final }}=\frac{u_{x \rightarrow j}^{\text {rec }} u_{i \rightarrow j}}{u_{i \rightarrow j}+u_{x \rightarrow j}^{\mathrm{rec}}-u_{x \rightarrow j}^{\text {rec }} u_{i \rightarrow j}} .
\end{array}\right.
\end{equation}
So the final reputation of $v_i$ to $v_j$ is
\begin{equation}
\mathcal{R}_{i \rightarrow j}^{\text {final }}=r_{i \rightarrow j}^{\text {final }}+\gamma u_{i \rightarrow j}^{\text {final }} .
\end{equation}

Reputation values are incorporated into the DAG as training model parameters. The DAG comprises nodes of both data sharing events and training model parameters, which are connected by edge connections established through approval relationships between transactions. These connections facilitate nodes to be connected to each other. Vehicles store DAGs locally and update asynchronously to achieve consistency with other vehicles. Each vehicle disperses its own latest DAG, which includes transactions and approval relationships, to neighboring vehicles. This scheme propagates updated DAGs and maintains loose consistency among vehicles, resulting in lower computational intensity and increased global robustness. The specific process is shown in Fig.~\ref{reputation}.
\subsection{Consensus: Proof of Reputation(PoR)}
By transforming the data sharing problem into a model sharing problem, the privacy of data holders is enhanced. Moreover, the data model of the graph neural network efficiently provides essential information for new sharing requests, as well as better-preserved vehicle trajectory features during data fusion. However, the use of prevailing consensus mechanisms like Proof-of-Work (PoW) in data sharing may lead to high computational and communication costs. To resolve this matter, implementing the FL Authorised Consensus Proof of Reputation (PoR) protocol is suggested. PoR combines vehicle reputation values with the consensus process, enhancing node computational resources. When data sharing requests occur, consensus committee members are chosen by retrieving the relevant blockchain nodes. The committee oversees the consensus process and acquires knowledge of the data model of the requested data. Asynchronous FL seeks to develop a worldwide information model $M$, which delivers an effective response $M$(Req) to inquiries concerning data sharing, thereby achieving the intended objective.

Sending consensus messages to only the committee nodes can lower communication overhead. However, reducing the number of nodes increases the challenge of reaching agreement. We offer reputation proof for consensus in data sharing to balance security and overhead. We choose committee leaders based on their reputation scores. As each committee node trains a local data model, it is imperative to verify and measure the quality of the model during the consensus process. Prediction accuracy is deemed essential to evaluate the performance of the trained local models. More precisely, we train the model as a regression task each time during trajectory prediction training. The total loss can be calculated as
\begin{equation}
\begin{aligned}
\text { Loss } & =\frac{1}{\mathcal{T}} \sum_{t=1}^{\mathcal{T}}\left\|Y_{p r e d}^t-Y_{G T}^t\right\|
\end{aligned}
\label{loss}
\end{equation}
where $\mathcal{T}$ is the future time step, and $Y_{p r e d}^t$ and $Y_{G T}^t$ are the predicted position and ground truth at time step $t$, respectively.

For each committee node, a trained global model $M$ and a local model $m_i$ are obtained after asynchronous training. The committee executes the consensus process, during which committee node $P_i$ sends its trained model $m_i$ to the next committee node while responding to a data sharing request. The transmission is recorded as a model transaction $t_{m_i}$ along with its corresponding loss ($m_i$). $P_i$ has a pair of public and private keys ($PK_i$, $SK_i$) to encrypt and sign the message. $P_i$ then sends the encrypted message $E\left(\mathrm{SK}_i\left(t_m\right), \mathrm{PK}_i\right)$ to the other committee nodes. All model transactions are then collected and stored locally as potential blocks by the committee node $P_j$.
As part of the training procedure, $P_j$ validates all transactions it receives by computing the loss defined in Eq.\ref{loss}. ${Loss}^u\left(P_j\right)$, the loss of $P_j$, is calculated through the following equation:
\begin{equation}
\operatorname{Loss}^u\left(P_j\right)=\gamma \cdot \operatorname{Loss}\left(M_j\right)+\frac{1}{n} \sum_i \operatorname{Loss}\left(m_i \right)
\end{equation}

where $\gamma$ represents a weight parameter indicating the contribution of $P_j$ to the global model and ${Loss}\left(M_j\right)$ represents the loss of the local training model $m_j$, determined by the size of the training data of $P_j$ and other participants $\gamma=1+\left|d_j\right| / \sum_i\left|d_i\right|$.

\subsection{Trajectory Prediction}
Using the aforementioned approach, we achieve effective interpretable federal learning for the trajectory prediction task. Specifically, by characterizing the vehicle trajectory data as a normalized weighted graph, the important private parameters are preserved. Preventing data redundancy, we achieve the elimination of vehicles with high similarity in the set of neighboring nodes using a multi-party query approach while protecting privacy. We are well aware that the quality of vehicle improvement data directly affects the results of parameter fusion in the federal learning model, so we introduce a reputation mechanism to solve this problem well by using the reputation values calculated from the private track data of the vehicles that are not eliminated and submit them to the reviewer for further grouping. The grouping method will be further described in section V.

\begin{figure*}[t!]
\centering
\setlength{\abovecaptionskip}{-0.1cm}
\includegraphics[width=6.5in]{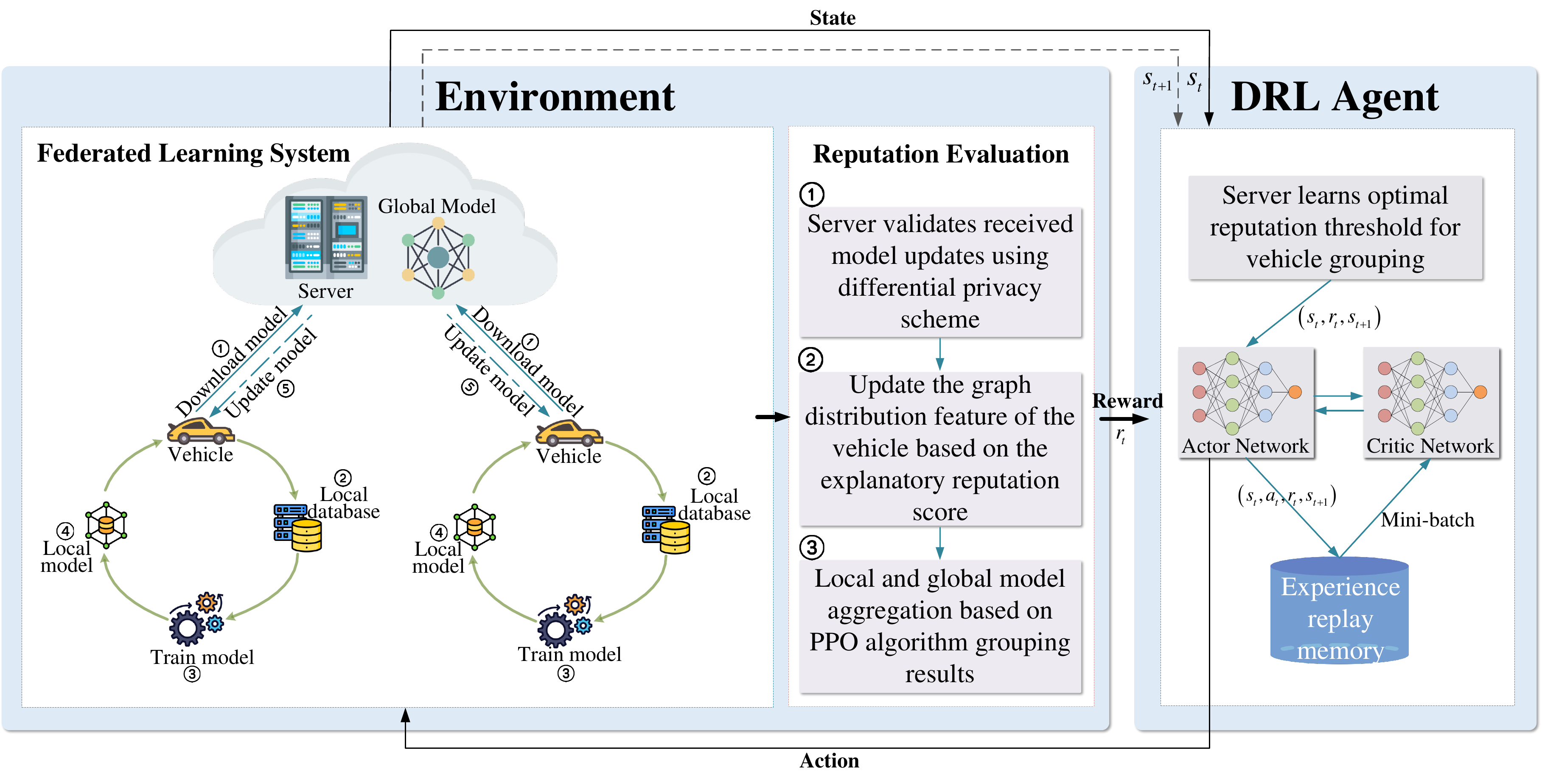}
\caption{Reputation-driven PPO vehicle clustering algorithm optimizes asynchronous federated learning to complete global model aggregation and improve vehicle local models.}
\label{PPO}
\vspace{-0.2cm}
\end{figure*}

\section{Enhanced Asynchronous Federated Learning with PPO Clustering Algorithm Based on Vehicle Reputation}
In this subsection, we tackle the challenges of insufficient heterogeneous model aggregation and significant global model update delays in FL by incorporating the proposed reputation evaluation mechanism to design an interpretable asynchronous FL mechanism that utilizes the enhanced PPO algorithm.

\subsection{Graph Asynchronous Federated Learning for Trajectory Prediction}
In IoV, the computational power and dynamic communication conditions of each vehicle result in different learning times. As a result, the slowest participant determines the running time of each learning iteration, while the other participants must wait to keep in sync. Existing work on trajectory prediction using FL essentially uses the synchronous federated algorithm FedAvg\cite{FedAVG}, which improves data security for multi-client model training. However, synchronous FL leads to increased communication costs and longer waiting times for idle nodes. For tasks with spatial timing dependencies, it can cause training difficulties leading to untimely global aggregation. Several studies have investigated asynchronous learning mechanisms to improve the learning performance\cite{asynchronousFL,ESWAAFL}. However, implementing these mechanisms for trajectory prediction may lead to suboptimal node selection, as their lack of interpretability can introduce biases in vehicle evaluations. Therefore, this paper proposes a method to predict trajectories via asynchronous FL. It utilizes the proposed reputation evaluation mechanism to group nodes with the PPO algorithm, enabling the asynchronous training of trusted vehicle clusters.

With the filtering and reputation value calculations of the previous section, we obtained a set of vehicles with unique features. Local aggregation on each vehicle $v_i$ is performed asynchronously within the range to enhance the quality of the locally trained model. The vehicles preserve and update the DAG locally to achieve asynchronous consistency with the other vehicle nodes. Asynchronous consensus enables vehicles to reach an agreement based on historical states rather than the current state. Vehicle $i$ sends updates to its local $D A G_i$ and then transmits the updates to nearby vehicles using a gossip scheme to achieve synchronization. Each vehicle randomly scatters its latest $D A G_i$ to neighboring vehicles. The gossip scheme propagates the DAG updates and maintains loose consistency between vehicles. This approach is less computationally intensive and globally robust. While asynchronous FL has improved aggregation efficiency, the issue of vehicle grouping remains unresolved. Limitations arise from exclusively conducting asynchronous local aggregation on nearby local vehicles to obtain a global model, as it fails to adapt to dynamic traffic conditions and thus cannot effectively optimize local prediction models. Therefore, as illustrated in Fig.~\ref{PPO}, we employ the PPO algorithm with an innovative reward feedback mechanism driven by reputation evaluation. This mechanism prioritizes the deep aggregation of high-reputation vehicles, leading to a superior global aggregation model.

\subsection{Reputation-Based PPO Vehicle Clustering Algorithm}
We aim to minimize execution time and enhance model accuracy by precisely categorizing the vehicle nodes according to their reputation values. During the global aggregation phase, the disparate computational resources and fluctuating communication conditions among different vehicles hinder the attainment of efficient execution. Therefore, we propose an approach where participating vehicles within a specific timestamp are selected based on reputation values until each grouping contains no more than $n$ vehicles. This approach prioritizes the aggregation of trajectory graphs with high reputation representing high data quality and improves the global aggregation.

Unlike conventional reinforcement learning, which evaluates the performance of an agent in a specific slot primarily based on accuracy loss alone, we consider the characteristics of the data during node clustering. For this purpose, we introduce the Reputation of Learning (RoL) to describe the high or low reputation value of vehicle $i$ in the aggregation process of slot $t$ as follows,
\begin{equation}
\mathcal{C}_r^t=\sum_{i \in V_P} \sigma_i^t\left(w^t, \mathfrak{d}_i\right)=\sum_{i \in V_P} \sum_j L\left(y_j-\frac{1}{\mathcal{R}^t\left(x_j\right)}\right),
\end{equation}
where $w^t$  represents the completed combined model in slot $t$, and $\mathfrak{d}_i=\left\{\left(x_j,y_j\right)\right\}$ represents the training data of vehicle $i$. Furthermore, we define the local learning time cost of vehicle $i$ in slot $t$ and the communication cost of vehicle $i$ as follows,
\begin{equation}
\mathcal{C}_a^t(i)=\frac{\mathfrak{d}_i \cdot \beta_m}{\xi_i(t)}, \mathcal{C}_u^t(i)=\frac{\left|w_i\right|}{\tau_i},
\end{equation}
where $\beta_m$ denotes the necessary CPU cycles to train model $m$ per iteration. Thus, the function of time consumption is
\begin{equation}
\mathcal{C}_{e}^t=\frac{1}{\left|V_P\right|} \sum_{i=1}^{\left|V_P\right|}\left(\mathcal{C}_a^t(i)+\mathcal{C}_u^t(i)\right).
\end{equation}
In this way, the total cost of FL in time slot $t$ can be given by the following equation:
\begin{equation}
\mathcal{C}_{f}^t\left(\lambda^t\right)={\mathcal{C}_r^t}+\mathcal{C}_{t e}^t,
\end{equation}
where $\lambda^t=\left[\lambda_i^t\right]$ in the time step $t$ is an indicator vector for vehicle selection, with $\lambda_i^t=1$ indicating activation and $\lambda_i^t=0$ indicating the opposite. We formulate the combinatorial optimization problem using a Markov Decision Process, which we denote as $\mathcal{M}=\left(S,V,P_v,C_v\right)$. The task of node selection can be expressed in the following way: 
\begin{subequations}\label{eq:2}
\begin{align}
\min _{\lambda^t} & \mathcal{C}_{f}^t\left(\lambda^t\right) \label{eq:2A}\\
\text { s.t. } & \lambda_i^t \in\{0,1\}, \quad \forall i, \label{eq:2B}\\
& \left|p_{i \mid \lambda_i=1}(t)-p_c(t)\right| \leq r_0^2, \label{eq:2c}
\end{align}
\end{subequations}
where the constraint Eq.\ref{eq:2c} ensures that the distance between the selected participating vehicles and the computed centroid cannot exceed a finite distance $r_0$. We use PPO to solve the problem Eq.\ref{eq:2A}. The fundamental principle involves updating the system policy with a value function.
\begin{algorithm}[!t]
\caption{The PPO Based Node Clustering Algorithm.}\label{alg:alg1}
\begin{algorithmic}
\STATE 
\STATE {\textsc{Input:} initial policy parameters $\theta_0$, initial value function parameters $\phi_0$}
\STATE \textbf{for} $k=0,1,2, \ldots$ \textbf{do}
\STATE \hspace{0.5cm}{Collect set of trajectories} $\mathcal{D}_k=\left\{\tau_i\right\}$ {by running policy} $\pi_k=\pi\left(\theta_k\right)$ {in the environment.}
\STATE \hspace{0.5cm}{Compute rewards-to-go} $\hat{R}_t$ 
\STATE \hspace{0.5cm}{Compute advantage estimates} $\hat{A}_t$, {(using any method of advantage estimation) based on the current value function $V_{\phi_k}$}
\STATE \hspace{0.5cm}Update the policy by maximizing the PPO objective:
$$
\begin{aligned}
& \theta_{k+1}=\arg \max _\theta \frac{1}{\left|\mathcal{D}_k\right| T} \sum_{\tau \in \mathcal{D}_k} \sum_{t=0}^T \min \\
& \left(\frac{\pi_\theta\left(a_t \mid s_t\right)}{\pi_{\theta_k}\left(a_t \mid s_t\right)} A^{\pi_{\theta_k}}\left(s_t, a_t\right), g\left(\epsilon, A^{\pi_{\theta_k}}\left(s_t, a_t\right)\right)\right)
\end{aligned}
$$
typically via stochastic gradient ascent with Adam.
\STATE \hspace{0.5cm}Fit value function by regression on mean-squared error:
$$
\phi_{k+1}=\arg \min _\phi \frac{1}{\left|\mathcal{D}_k\right| T} \sum_{\tau \in \mathcal{D}_k} \sum_{t=0}^T\left(V_\phi\left(s_t\right)-\hat{R}_t\right)^2
$$
typically via some gradient descent algorithm.
\STATE \textbf{end for}
\end{algorithmic}
\label{alg1}
\end{algorithm}
We utilize PPO to ascertain the optimal resolution for vehicle reputation categorization in asynchronous FL. PPO operates on a well-defined Markov decision process $\mathcal{M}=\left(S,V,P_v,C_v\right)$. At each time slot $t$ in FL, the system state is represented by $s_t=\{\boldsymbol{\tau}(t), \boldsymbol{\xi}(t), \boldsymbol{\gamma}(t), \boldsymbol{\lambda}(t-1)\}$. Here, $\boldsymbol{\tau}(t)$ denotes the wireless data rate between vehicles, $\boldsymbol{\xi}(t)$ represents the available computing resources of the vehicles, $\boldsymbol{\gamma}(t)$ indicates the reputation of the vehicles, and $\boldsymbol{\lambda}(t-1)$ signifies the selection state of the vehicles. The action taken at time slot $t$, denoted by $\lambda^t=\left(\lambda_1^t, \lambda_2^t, \ldots, \lambda_n^t\right)$, corresponds to a vehicle selection decision and can be framed as a 0-1 problem. Specifically, $\lambda_i^t = 1$ when vehicle $i$ is selected as a node with a high reputation value, otherwise $\lambda_i^t = 0$. Our improvement is mainly in the reward function. Specifically, we customize the reward function to add 1 to the reward value for choosing a vehicle with a high reputation value, subtract 1 from the reward value for choosing a low reputation value, subtract 10 from the reward value for choosing a low reputation value for 5 consecutive times, and end the round; if the number of vehicles chosen with a high reputation value reaches the threshold, then 20 points are rewarded; and end the round. The system assesses the impact of an action using a reward function called $R$. At iteration $t$, the agent carrying out the task of selecting the vehicle reputation takes action $\lambda_t$ while in state $s_t$. Subsequently, the action is appraised via the established reward function in the ensuing manner:
\begin{equation}
\begin{aligned}
R\left(s_t, \lambda_t\right)=- & \frac{1}{\left|\sum_{i=1}^n \lambda_i\right|} \sum_{i=1}^n \mathcal{C}_i^t \cdot \lambda_i^t \\
=-\frac{1}{\left|\sum_{i=1}^n \lambda_i\right|} & \left(\sum_{i=1}^n \lambda_i\left(\frac{\mathfrak{d}_i \cdot \beta_m}{\xi_i(t)}+\frac{\left|w_i\right|}{\tau_i}\right)\right. \\
 & \left.+\sum_{i=1}^n \lambda_i \sigma_i^t\left(w^t, \mathfrak{d}_i\right)\right)
\end{aligned}
\end{equation}
The reward function $R\left(s_t, \lambda_t\right)$ assesses the standing of taking action $\lambda_t$ at time interval $t$. The overall cumulative reward can be presented as: 
\begin{equation}
\mathbb{E}\left[\sum_{t=0}^{T-1} \gamma R\left(s_t, \lambda_t\right)\right]
\end{equation}
where $\gamma \in(0,1]$ is the reward discount factor.
For the PPO algorithm, the objective is to determine the value of lambda that maximizes reputation accumulation, as shown in the following Eq.\ref{argmax}.
\begin{equation}
\label{argmax}
\lambda=\arg \max \mathbb{E}\left[\sum_{t=0}^{T-1} \gamma R\left(s_t, \lambda_t\right)\right]
\end{equation}

Through algorithm \ref{alg1}, the corresponding groups with high and low reputations can be obtained. Next, the vehicle feature parameters of each group are combined, followed by global fusion, with priority given to the high reputation group for deep aggregation.

\begin{table}[t!]
\centering
\caption{Dataset profile.}
\begin{tabular}{lll}
\hline
Attributes               & ApolloScape        & NGSIM   \\ \hline
Frequency of sample      & 2 Hz (downsampled) & 5 Hz    \\
Total number of vehicles & 5156               & 11 779  \\
Total frames             & 5593 (downsampled) & 30 476  \\
Train len/pred len       & 3 s/3 s            & 3 s/5 s \\ \hline
\end{tabular}
\label{dataset}
\end{table}

\section{EXPERIMENTS}
In this section, we initially present the renowned datasets used in the experiment, followed by the description of evaluation metrics. Subsequently, we explicate the impact of adjusting diverse model components on the prediction effect and compare the current approach to previous ones. We investigate the impact of various iterations of the model on convergence and the influence of varying quantities of bad nodes on model forecasts. We run our scheme on a desktop running Ubuntu 20.04 with 2.30GHz Intel(R) Xeon(R) E5-2686 v4 CPU, 64GB Memory, and a NVIDIA 3090 Ti Graphics Card.

\begin{figure}[!t]
\centering
\setlength{\abovecaptionskip}{-0.02cm}
\includegraphics[width=3.3in]{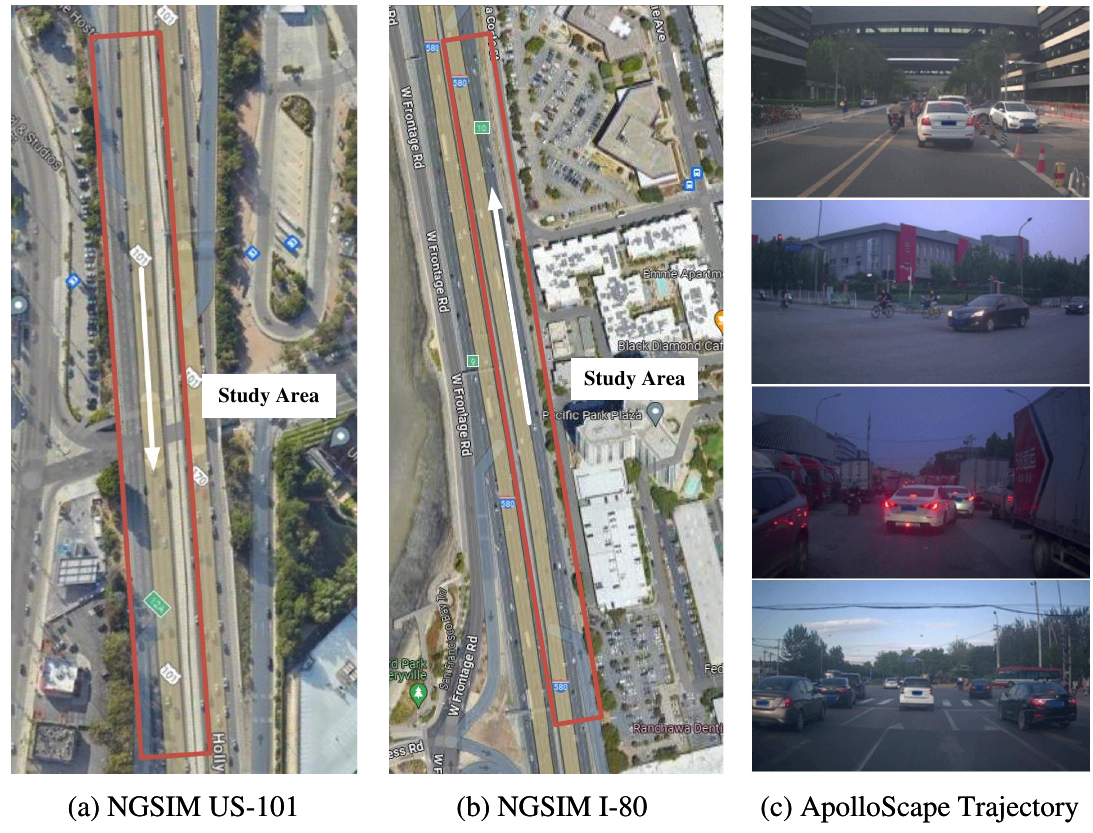}
\caption{The datasets used in the experiments: The vehicle trajectory data in the NGSIM dataset are collected by multiple digital cameras and include different traffic conditions such as light, moderate, and heavy congestion on real highways. Overhead views of the two study areas, US-101 highway \cite{U101} and I-80 highway \cite{I80}, are shown in Fig. (a) and (b), respectively, obtained from Google Maps. The ApolloScape trajectory dataset \cite{Apollo} consists of data collected by a vehicle named "Apollo acquisition car" during rush hours in urban areas. Our proposed approach is validated in these two different traffic environments.}
\label{data}
\vspace{-0.5cm}
\end{figure}

\subsection{Datasets}
We evaluated our proposed method RAFLTP on well-known real-world datasets: NGSIM\cite{I80,U101}, and ApolloScape Trajectory dataset\cite{Apollo}, as shown in Fig.~\ref{data}. The basic info is listed in Table~\ref{dataset}. 
NGSIM contains two road segment datasets, US-101 and I-80. Referring to the method\cite{Deo_How,DeoMulti,CS-LSTM} for data segmentation of the training set and testing set, a quarter of each of the three road condition data is selected as the test set. Each trajectory is segmented into 8 s, the first 3 s are used as training data, and the last 5 s are used as training labels.
The ApolloScape Trajectory dataset was gathered in an urban area during rush hour by a vehicle called the Apollo acquisition car. The data primarily comprises vehicle trajectories that are based on object detection and tracking algorithms. During the initial phase, we adopt GRIP++\cite{grip2} to select 20$\%$ of the sequences for validation purposes, while the remaining 80$\%$ is utilized as the training set.

\subsection{Metrics}
This subsection highlights several established metrics that demonstrate the validity and merits of the proposed methodology. Specifically, we adopt the Root Mean Square Error (RMSE), Average Displacement Error (ADE), and Final Displacement Error (FDE) to evaluate the performance with reference to Social-GAN\cite{SocialGAN}. 
\begin{figure}[t]
\centering
\setlength{\abovecaptionskip}{-0.02cm}
\includegraphics[width=2.8in]{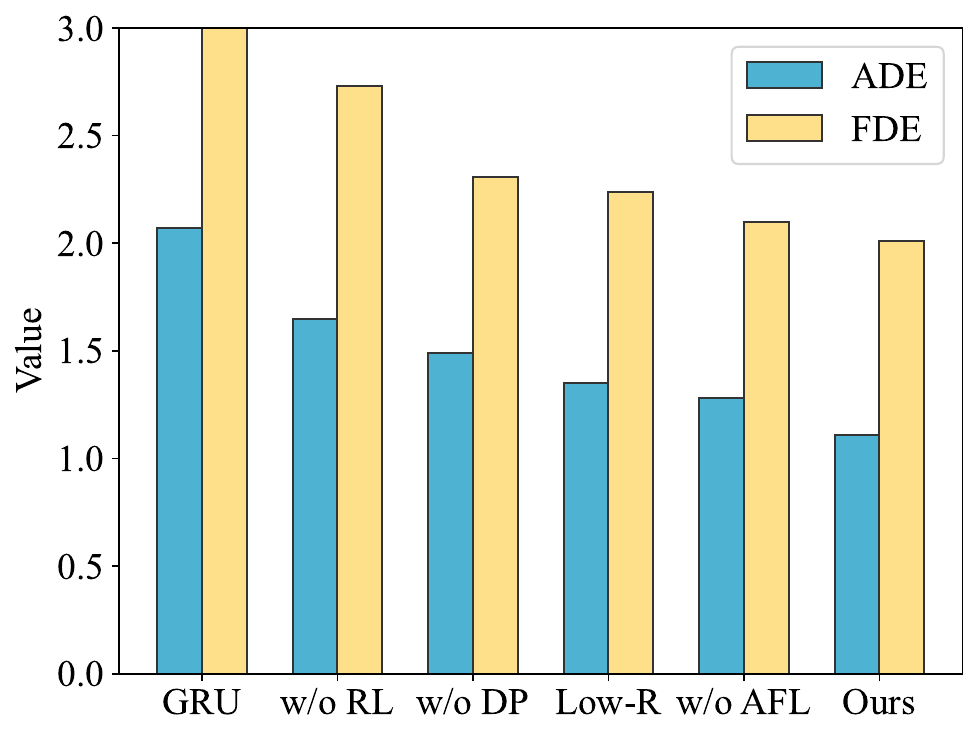}
\caption{The influence of different components.}
\label{abla}
\end{figure}



\subsection{Ablation study}
To demonstrate the advantage of the proposed approach, several subsequent ablation experiments were performed in this subsection.
\begin{figure}[!t]
\centering
\setlength{\abovecaptionskip}{-0.02cm}
\includegraphics[width=2.9in]{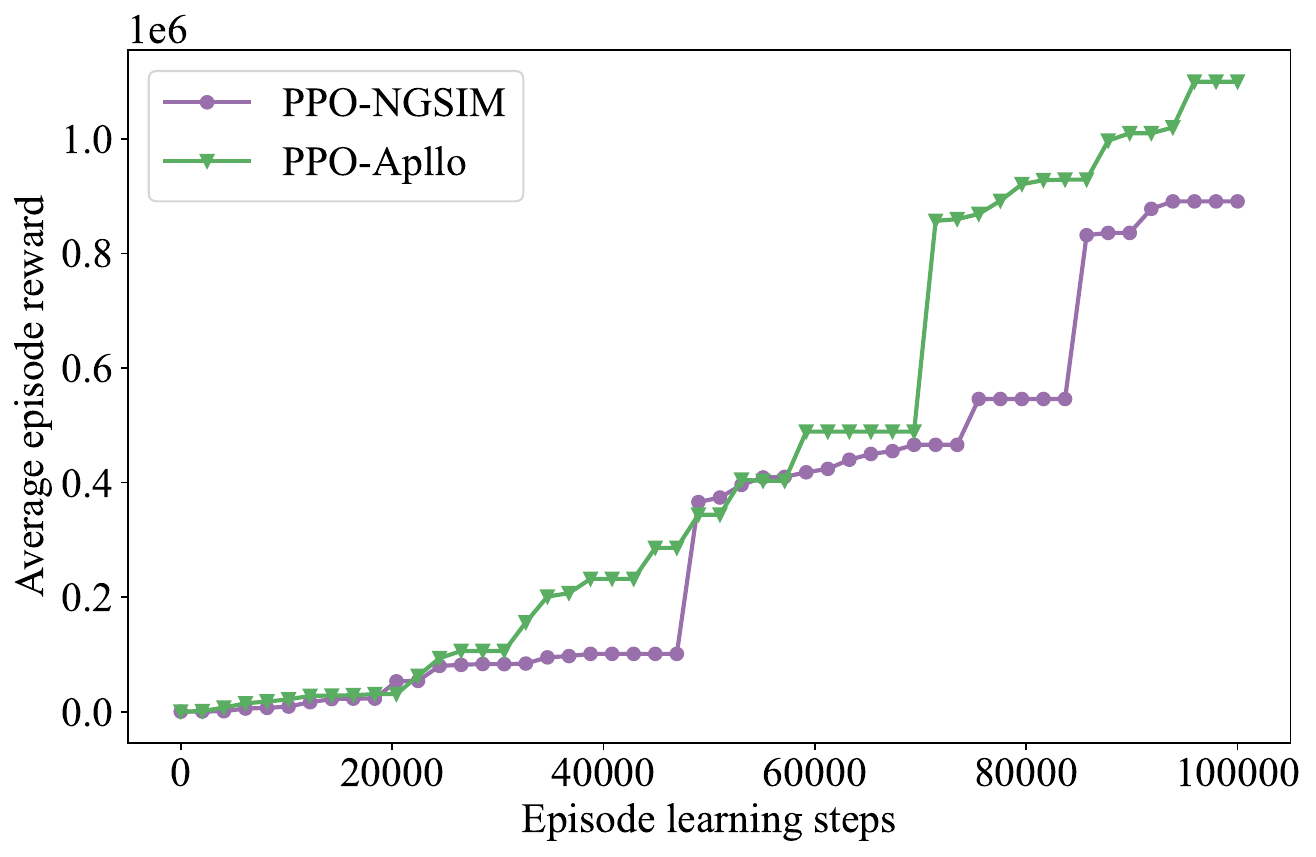}
\caption{Average accumulated episode reward of the PPO algorithm during training on different datasets.}
\label{ablappo}
\end{figure}

\begin{figure}[!t]
\setlength{\abovecaptionskip}{-0.02cm}
\centering
\includegraphics[width=3in]{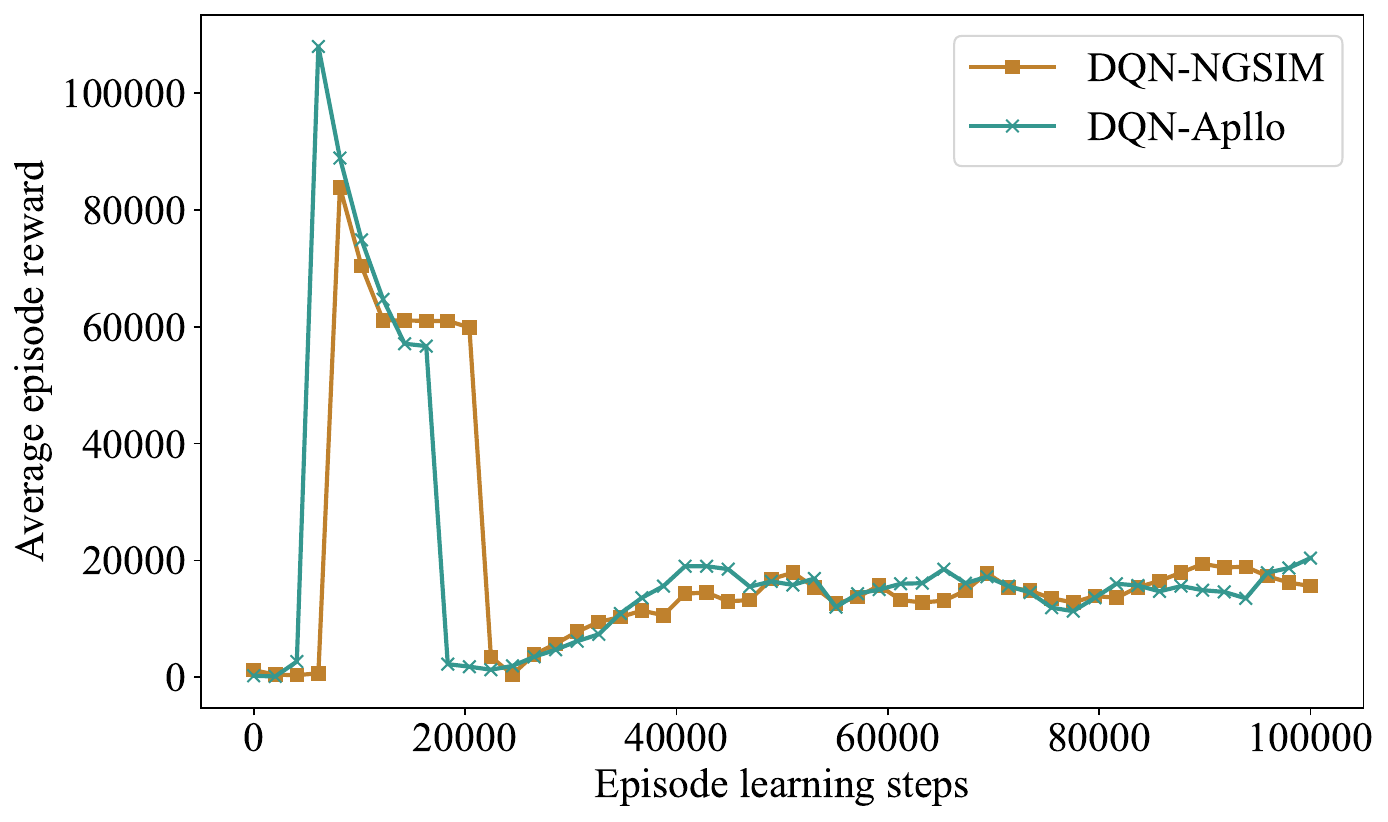}
\caption{Average accumulated episode reward of the DQN algorithm during training on different datasets.}
\label{abladqn}
\vspace{-0.3cm}
\end{figure}
\subsubsection{The influence of different components}
To test the effectiveness of the various components in our framework, we conducted ablation experiments on the ApolloScape Trajectory dataset consisting of five variants: basic encoder–decoder network (GRU based),  w/o RL, w/o differential privacy, w/o Asynchronous FL (AFL), priority aggregation of low reputation vehicles (Low-R) and our complete methodology. From Fig.~\ref{abla}, it is evident that the basic encoder-decoder network exhibits the highest error, primarily due to its limited consideration of vehicle interactions. However, with the incorporation of GRIP-based model enhancements under our proposed method, the ADE shows a substantial reduction. Furthermore, we observe that the impact of our method on ADE is most pronounced when the reputation-based PPO module is absent. This underscores the significance of our reputation-based PPO node grouping approach in trajectory prediction. By dynamically aggregating data from high-quality vehicle data providers using learnable weights, we effectively enhance prediction accuracy.

Additionally, the influence of the differential privacy module is apparent, as it enforces constraints to filter out abnormal and redundant data, resulting in more robust and secure model aggregation. It's noteworthy that the absence of FL leads to lower prediction accuracy compared to the full model. This emphasizes the effectiveness of reputation-based grouping aggregation in alleviating the challenges associated with heterogeneous data fusion and enhancing trajectory prediction accuracy. Moreover, if we choose to prioritize the aggregation of vehicles with low reputation values, it will rather reduce the trajectory prediction accuracy.

\subsubsection{The influence of DRL Methods}
We compare the learning results of the PPO algorithm and the DQN algorithm in the NGSIM and ApolloScape datasets. All algorithms take the same number of network layers, neurons, and training parameters. Fig. \ref{ablappo} and Fig. \ref{abladqn} demonstrate that the PPO algorithm is more stable and adaptable to changes in both datasets, converging quickly after positive feedback. Additionally, it was found through testing that the training time of the PPO algorithm is 34.5\% faster than that of the DQN algorithm.

\begin{figure}[t!]
\centering
\includegraphics[width=3in]{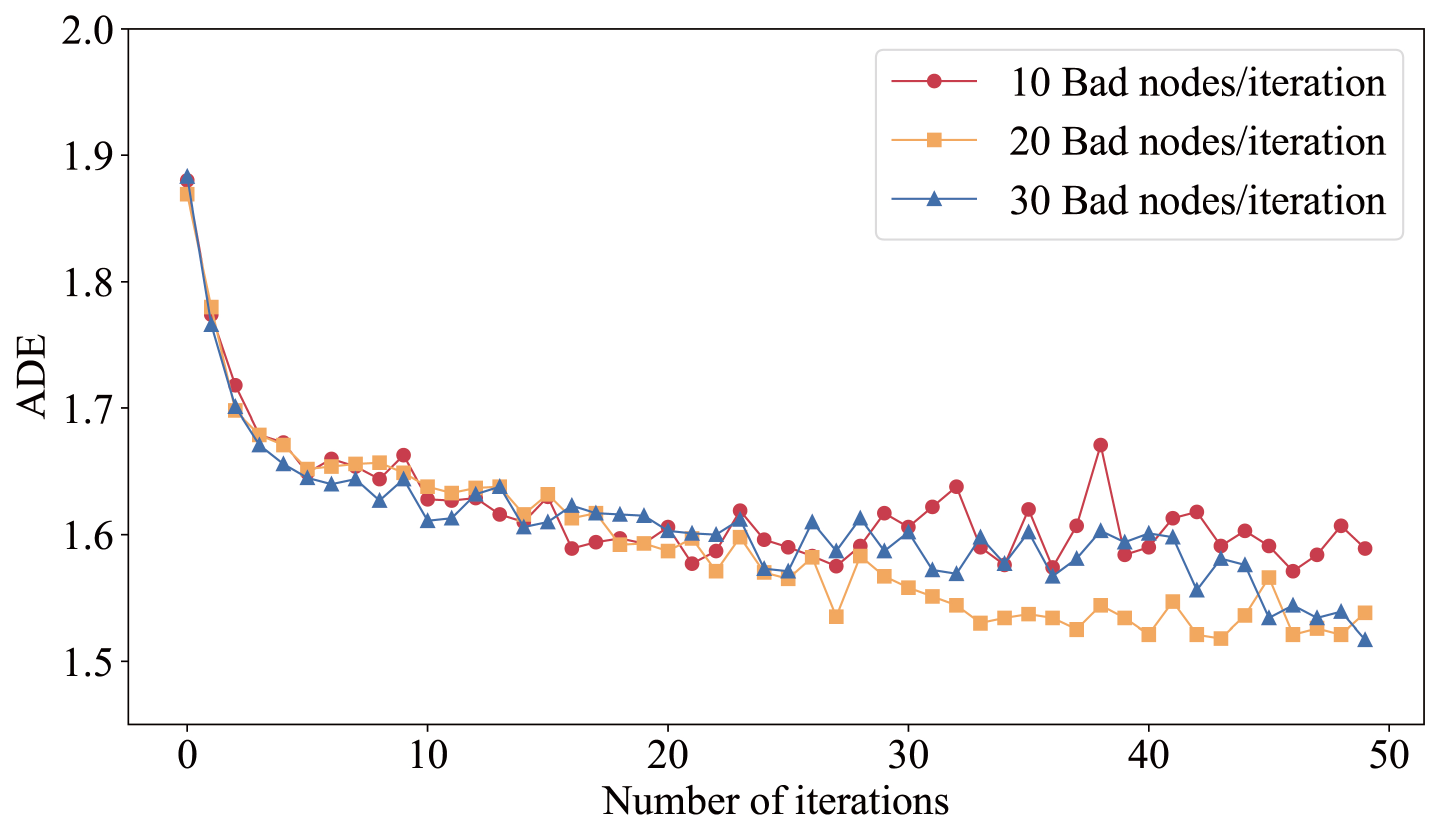}
\caption{The impact of Bad Nodes on ADE w/o Differential Privacy}
\label{badnode_NoDP}
\end{figure}

\begin{figure}[t!]
\centering
\includegraphics[width=3in]{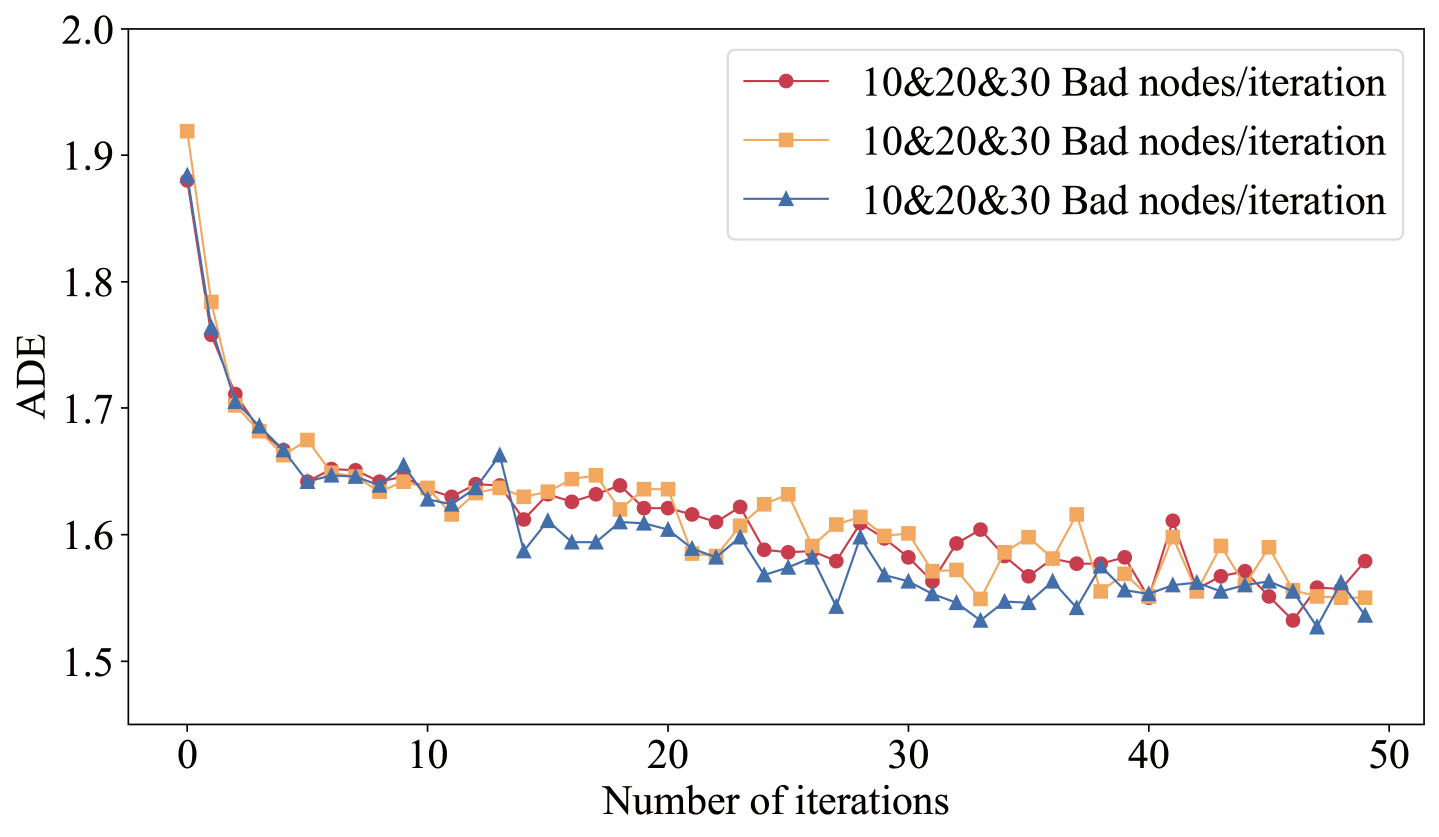}
\caption{Impact of Bad Nodes on ADE: Our Model (Blue Solid Line) vs. w/o Reinforcement Learning (Orange Solid Line) vs. w/o FL Module (Red Solid Line)}
\label{badnode_NoFRA}
\end{figure}

\subsubsection{The influence of bad nodes}
To analyze the impact of bad nodes on different modules within the overall framework, we examined how different components of the proposed solution affect the ADE under different amounts of bad nodes. We randomly selected a certain number of data providers at each iteration and manipulated their data to introduce anomalies.
As shown in Fig.~\ref{badnode_NoDP} and Fig.~\ref{badnode_NoFRA}, an increase in the number of bad nodes significantly degrades performance. Interestingly, in Fig.~\ref{badnode_NoDP}, the w/o differential privacy leads to different effects of ADE with different amounts of bad nodes. Surprisingly, a higher number of bad nodes results in a lower ADE. This can be attributed to the reinforcement learning-based grouping of vehicles according to their reputation scores, which directs low-quality data providers to a shallow aggregation group and subsequently improves prediction accuracy.

In Fig.~\ref{badnode_NoFRA}, we can clearly observe the importance of the differential privacy module in safeguarding the model from bad nodes. The number of bad nodes has no discernible effect on the ADE of trajectory prediction, confirming the efficiency of the differential privacy module in constraining the data structures shared by the data providers. This effectively limits the impact of anomalous data nodes and redundant nodes on prediction accuracy.


\begin{figure}[t!]
\centering
\includegraphics[width=3in]{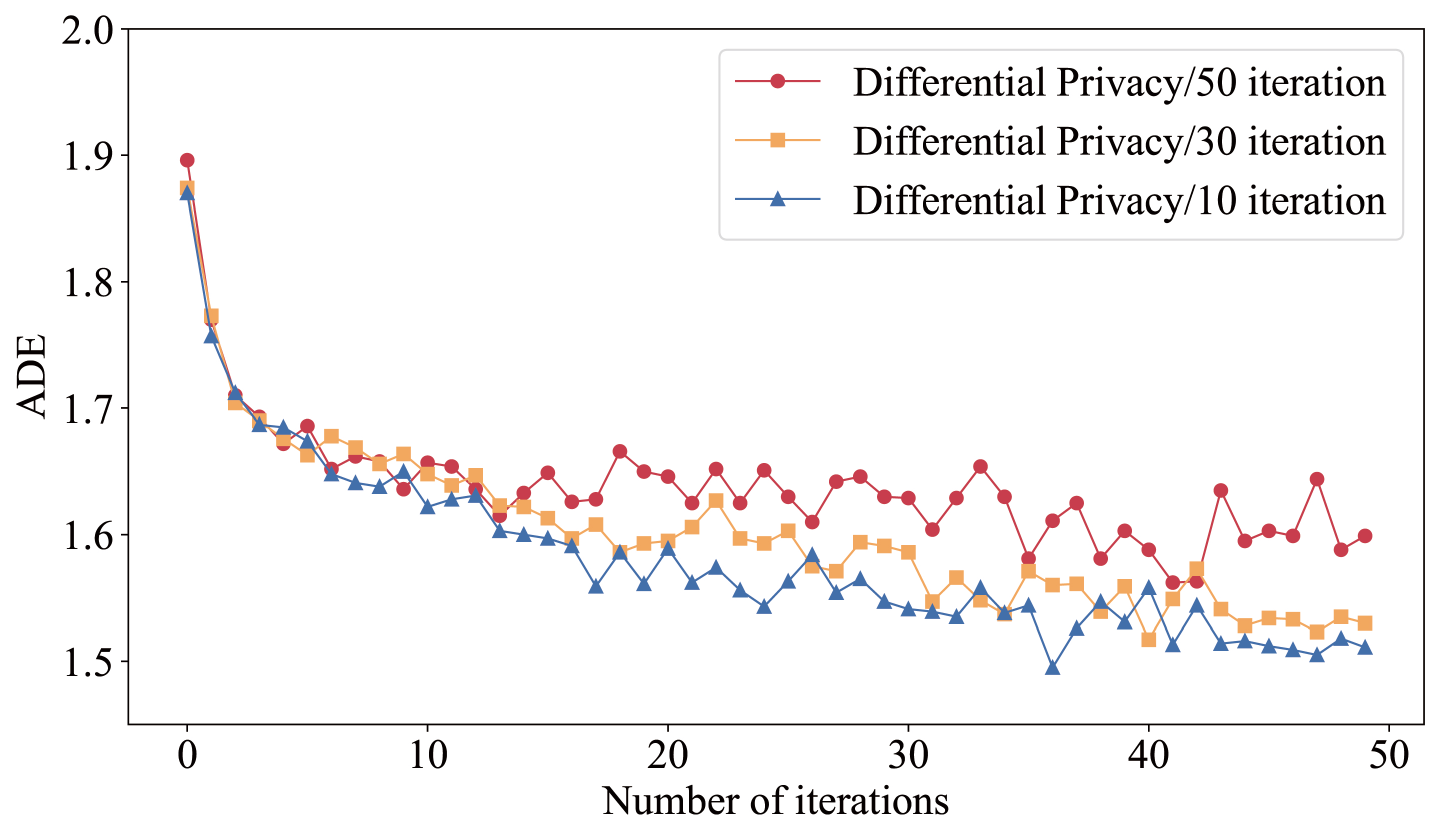}
\caption{Effect of Differential Privacy module invocation frequency on ADE.}
\label{FOARL}
\end{figure}
\begin{figure}[t!]
\centering
\includegraphics[width=3in]{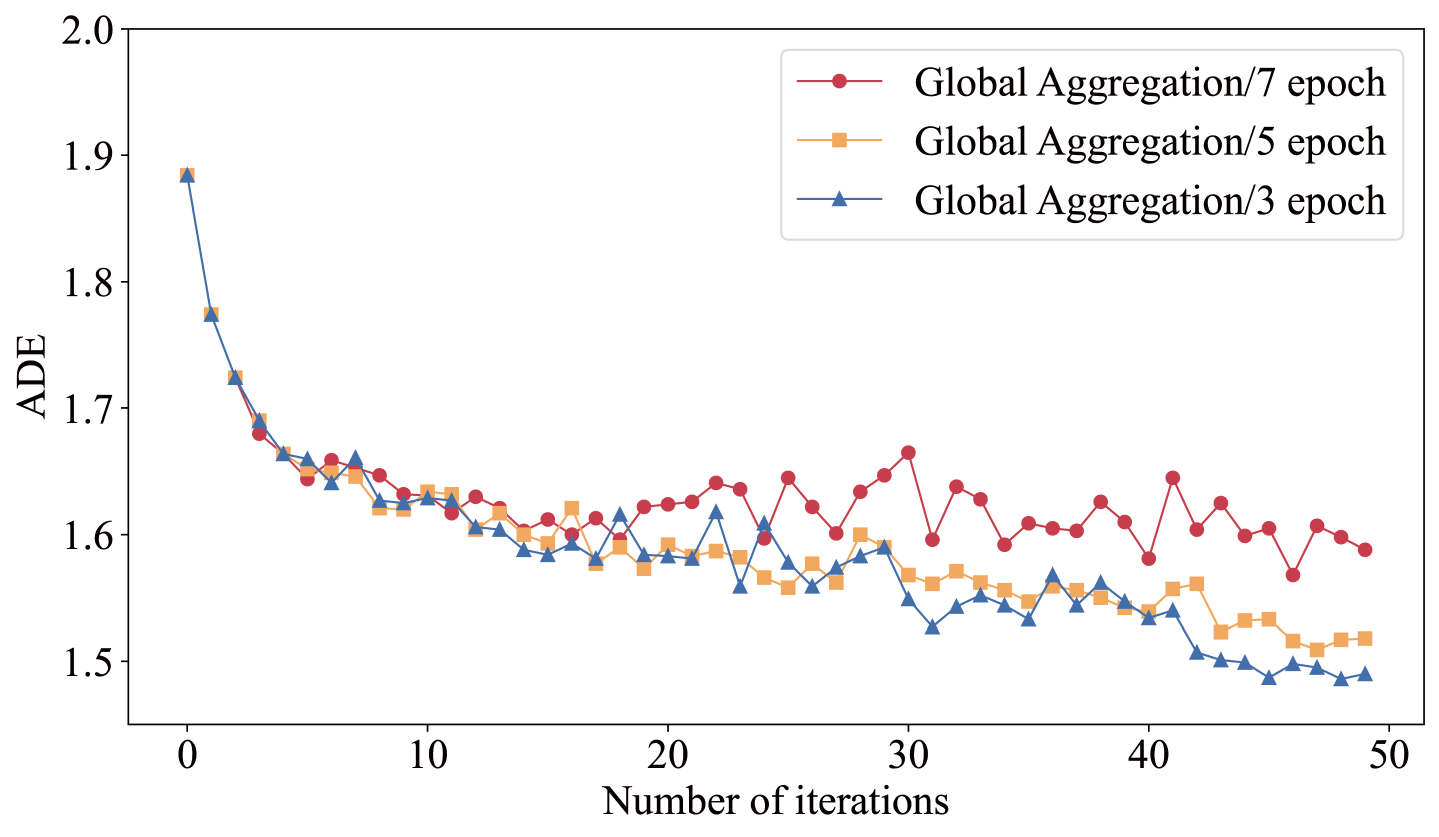}
\caption{Effect of global aggregation frequency on ADE.}
\label{FOADP}
\end{figure}
\subsubsection{Frequency of aggregation}
To explore the impact of different local and global aggregation frequencies on prediction accuracy and the influence of invoking the differential privacy module at different frequencies, we designed different combinations of aggregation and invocation frequencies. As depicted in Fig.~\ref{FOARL}, higher local and global aggregation frequencies lead to faster convergence and improved prediction accuracy. Similarly, Fig.~\ref{FOADP} shows that higher invocation frequencies of the differential privacy module also lead to improved prediction accuracy. Taken together, these results underscore the effectiveness of our proposed solution in improving prediction accuracy.


\subsubsection{Impact of Private Budget}
\begin{figure}[!t]
\centering
\setlength{\abovecaptionskip}{-0.03cm}
\includegraphics[width=2.8in]{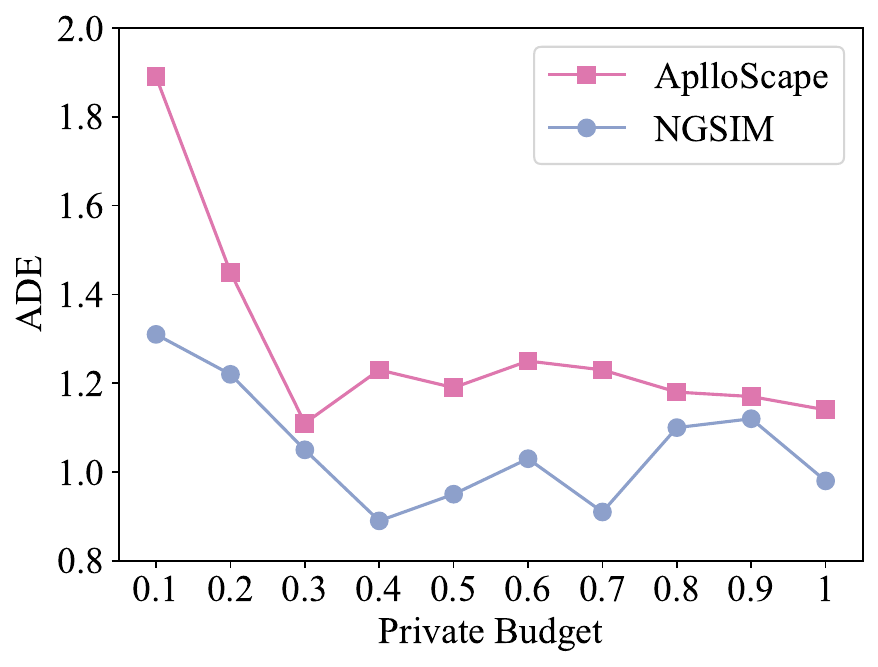}
\caption{Impact of Private Budget $\epsilon$ on ADE in different datasets.}
\label{budget}
\vspace{-0.3cm}
\end{figure}

To explore the impact of differential privacy hyperparameters on prediction accuracy while striking a balance between data utility and privacy, we investigate privacy budget $\epsilon$ from 0.1 to 1.0 across different datasets. As shown in Fig.~\ref{budget}, in the ApolloScape dataset, $\epsilon<0.3$ significantly affects data utility due to excessive noise, with stability observed from $\epsilon = $ 0.3 onwards. However, excessively large values of $\epsilon$ compromise privacy protection. Therefore, $\epsilon =$ 0.3 strikes a good balance between data privacy and utility.  In the NGSIM dataset, due to the increase in the number of vehicles, maintaining a balance between privacy budget and utility requires $\epsilon = 0.4$.
\begin{figure}[!t]
\centering
\setlength{\abovecaptionskip}{-0.005cm}
\includegraphics[width=2.8in]{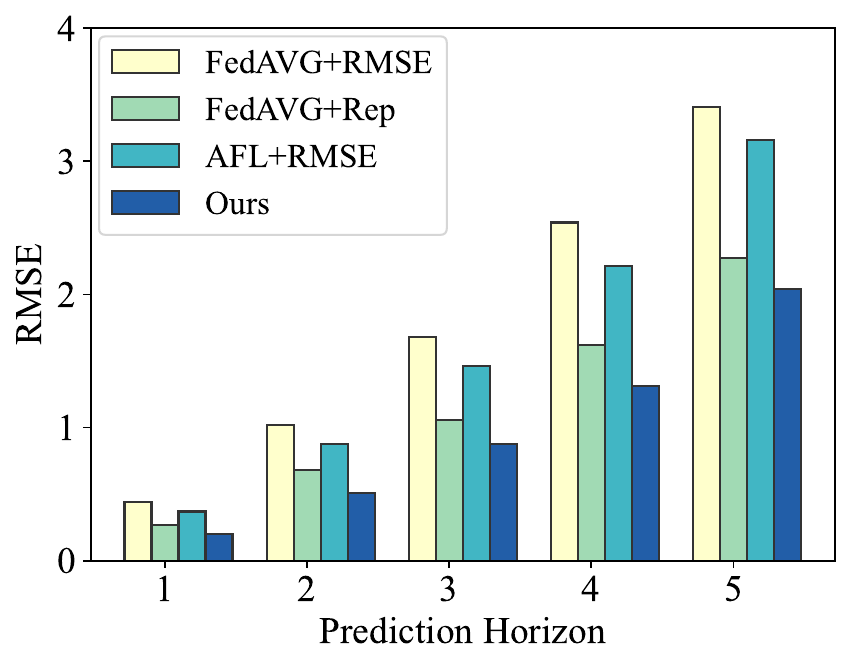}
\caption{Impact of reputation quantization mechanism.}
\label{fed}
\vspace{-0.3cm}
\end{figure}
\subsubsection{Impact of Reputation Quantization Mechanism}
To demonstrate the superiority of our proposed interpretable reputation reward mechanism as a quantitative metric for optimizing federated learning, we conducted ablation experiments on the NGSIM dataset. Specifically, we replaced our AFL algorithm with the traditional synchronous FL (SFL) scheme FedAVG \cite{FedAVG} and used RMSE as the trajectory prediction loss for uploading local parameters to the server for vehicle grouping. As depicted in Fig.~\ref{fed}, the traditional SFL scheme yielded inferior prediction results compared to the AFL scheme. In dynamic trajectory prediction scenarios, waiting for inefficient vehicle nodes to complete local model aggregation inevitably leads to suboptimal global model outcomes. Additionally, substituting RMSE for reputation values for reinforcement learning-based vehicle grouping resulted in a reduction in trajectory prediction accuracy. This is because loss functions provide an incomplete assessment of the impact of vehicle nodes on both local and global aspects. In dynamic traffic environments, vehicles continuously influence each other, leading to unavoidable biases in trajectory prediction. Discarding vehicle nodes solely based on significant biases and rewarding vehicles that consistently exhibit smooth and straight trajectories would be unfair. In contrast, our proposed vehicle reputation quantification mechanism focuses on the quality of vehicle data and the similarity of local trajectories. It facilitates deep aggregation of clusters of vehicles with high-quality data and similar trajectory model structure, thereby yielding a global model with enhanced guidance value.
\begin{figure*}[t!]
\centering
\subfloat[]
{\includegraphics[width=1.25in]{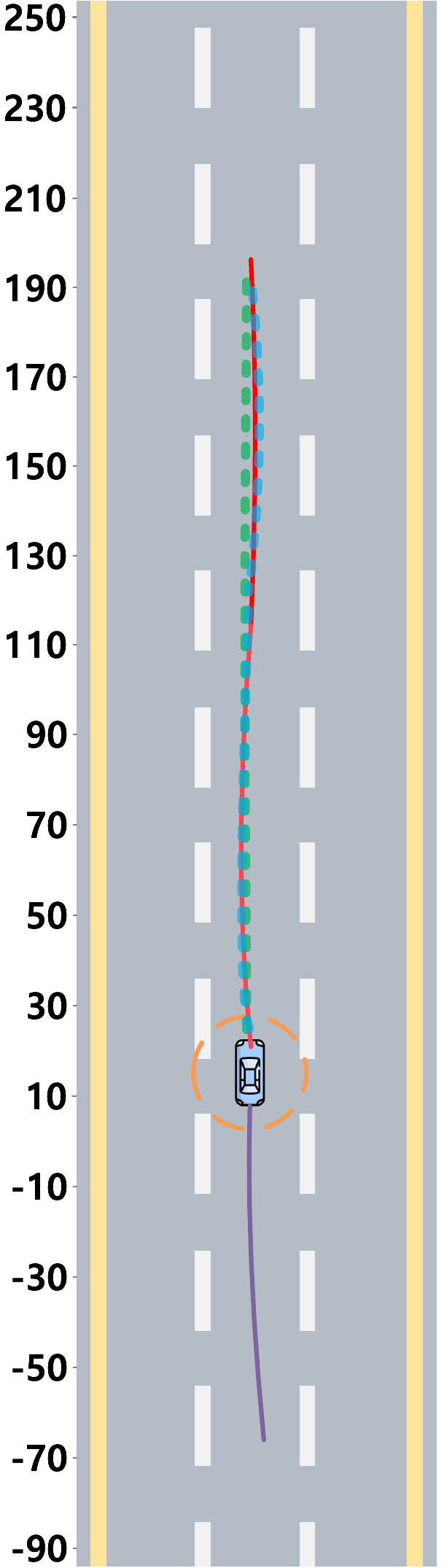}%
\label{visual:1}}
\hfil
\subfloat[]
{\includegraphics[width=1.25in]{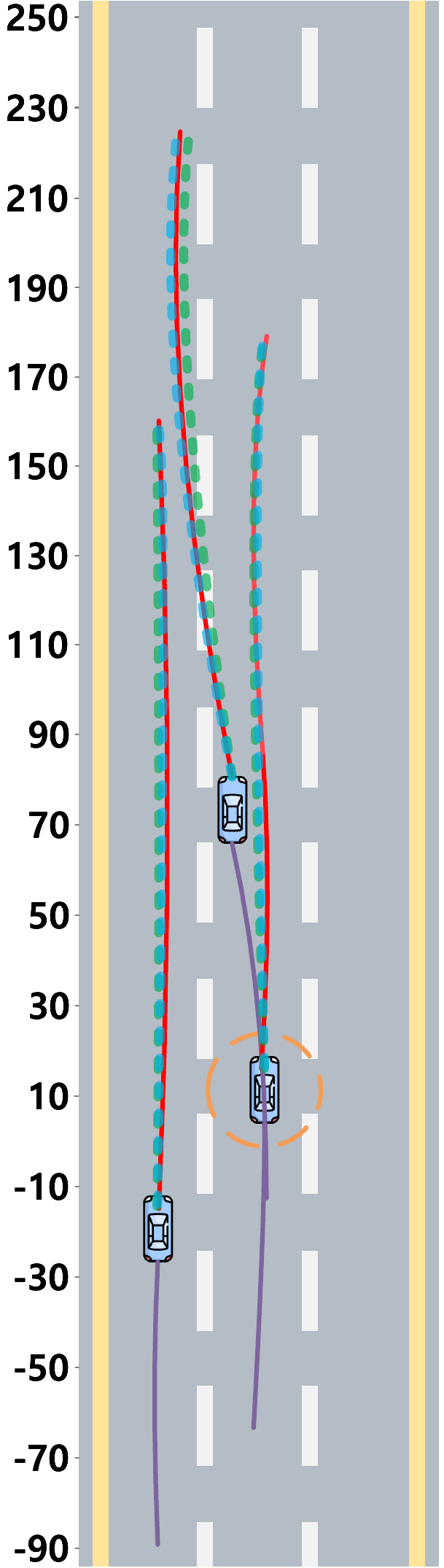}%
\label{visual:2}}
\hfil
\subfloat[]
{\includegraphics[width=1.25in]{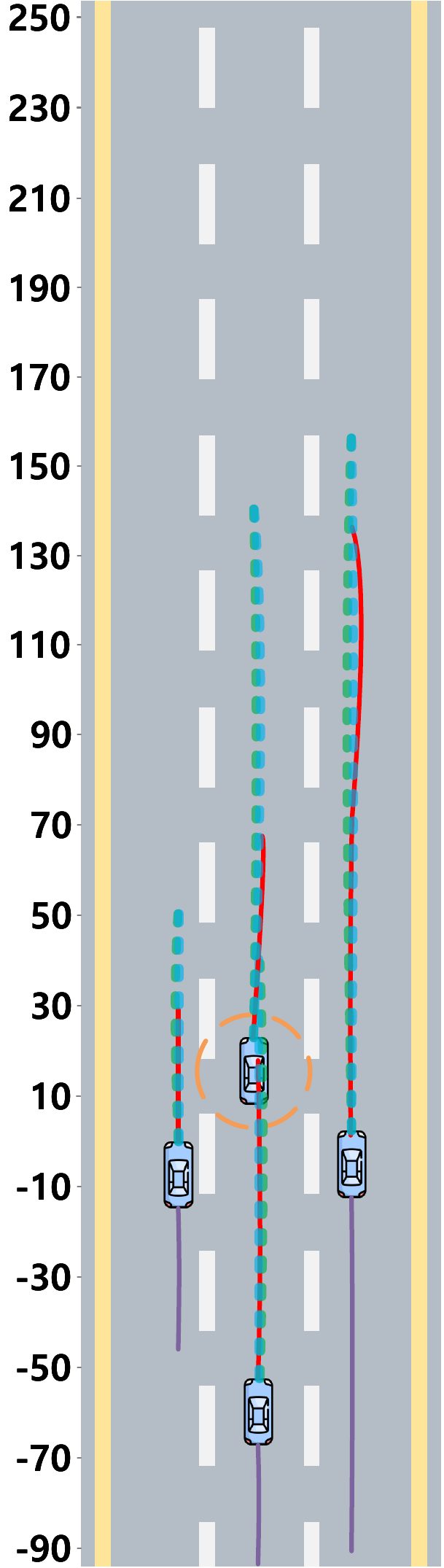}%
\label{visual:3}}
\hfil
\subfloat[]
{\includegraphics[width=1.25in]{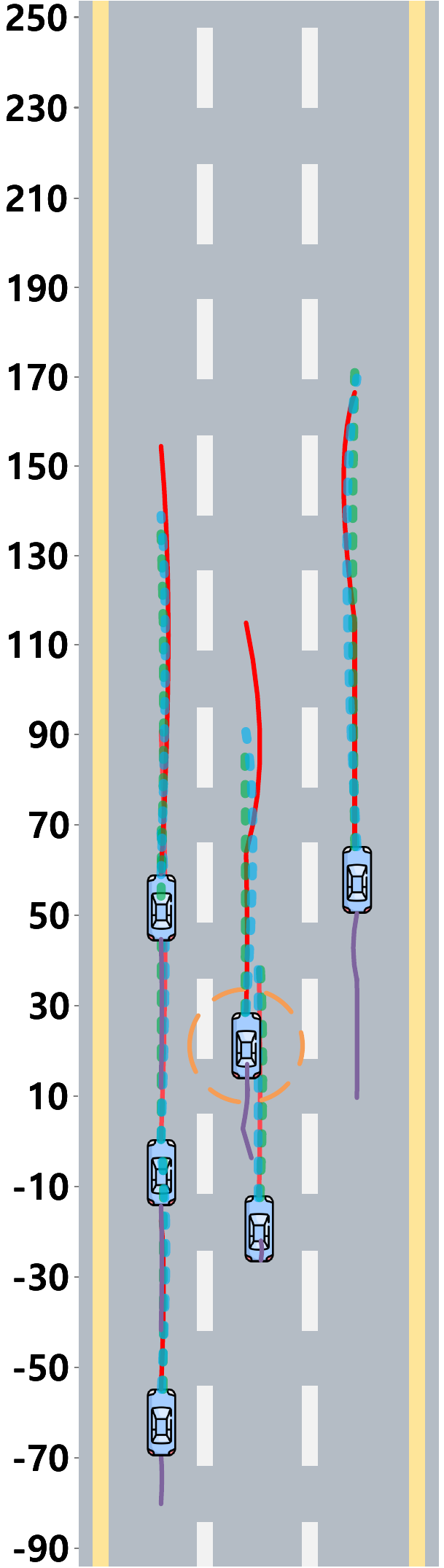}%
\label{visual:4}}
\hfil
\subfloat[]
{\includegraphics[width=1.25in]{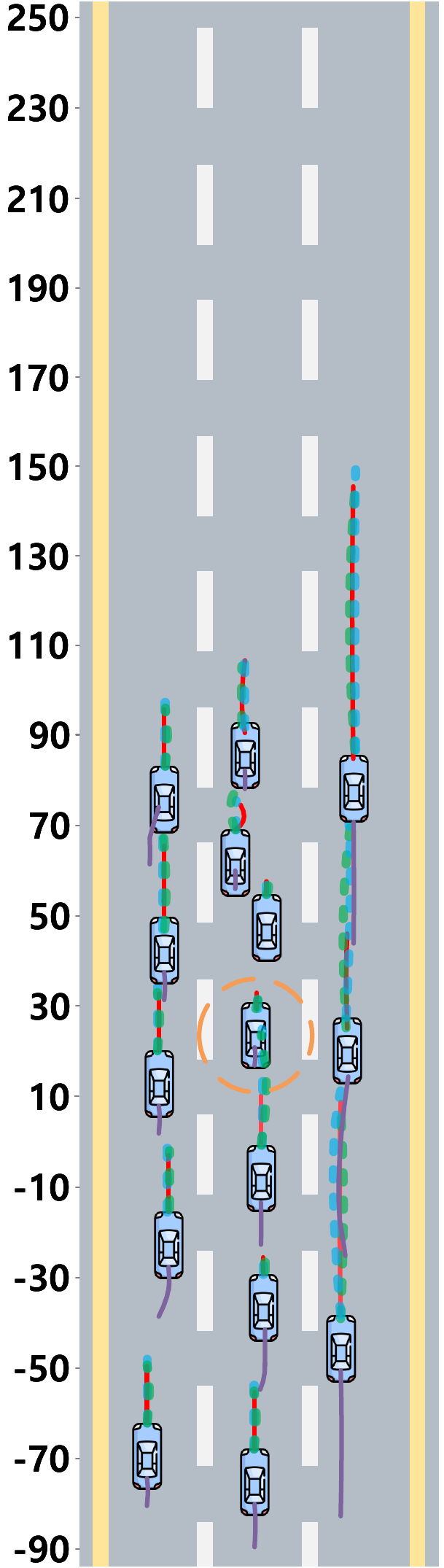}%
\label{visual:5}}
\hfil
\caption{Visualization of predicted trajectories under mild, moderate, and congested traffic scenarios, where the circled vehicles are those that GRIP++ tries to predict, the purple solid line is the observed history, the red solid line is the future ground truth, the blue dashed line is the prediction result of our model (5 seconds), the green dashed line is the GRIP++ prediction (5 seconds). (a) only one vehicle; (b) two vehicles with few interactions; (c) three vehicles with more interactions; (d) five vehicles with some interactions in which a vehicle would conduct lane changing; (e) heavy traffic.}
\label{viusal}
\end{figure*}

\subsection{Comparison analysis}
With regard to the latest findings in DGInet\cite{DGInet}, we have compared our model with seven baseline methods consisting of state-of-the-art trajectory prediction methods:
\begin{enumerate}
\item{Encoder–Decoder\cite{Spectral}: This approach employs an enc-dec architecture based on LSTM.}
\item{CS-LSTM\cite{CS-LSTM}: This approach combines CNN and LSTM to extract spatio-temporal features.}
\item{TraPHic\cite{TraPHic}: This approach employs spatial attention pooling in combination with CNNs and LSTMs to predict the trajectory.}
\item{Social-GAN\cite{SocialGAN}: It employs the encoder-decoder architecture as a generator and subsequently trains an additional encoder to serve as a discriminator for trajectory prediction.}
\item{GRIP\cite{GRIP}: This method uses graph convolution for trajectory prediction. The resulting features from the enc-dec network facilitate trajectory prediction.}
\item{Spectral\cite{Spectral}: The approach forecasts both paths and actions and utilizes an additional network founded on spectral clustering to adapt the LSTM-based enc-dec network.}
\item{DGInet\cite{DGInet}: This approach combines a semi-global graph mechanism with a convolutional graph network based on M-products for predicting trajectories.}
\end{enumerate}

In Table~\ref{perform}, we compare the ADE and FDE of our predicted trajectory with the previous methods. Unfortunately, our solution did not achieve state-of-the-art results. However, compared to the GRIP method, we managed to reduce the ADE by 11.2\% and the FDE by 10.2\% on the ApolloScape dataset, surpassing even the Spectral result. It's important to note that our approaches, like GRIP and DGInet, predict trajectories for all vehicles in a traffic scenario, as opposed to Spectral, which predicts only a single vehicle's trajectory. Fortunately, our approach, which enhances the security and robustness of data sharing among vehicles in a distributed trajectory prediction scenario, aligns more closely with real-world traffic situations and stands out as a unique contribution.

\begin{table}[!t]
\centering
\caption{AVERAGE PERFORMANCE COMPARISON.}
\begin{tabular}{lllll}
\hline
\multirow{2}{*}{Method} & \multicolumn{2}{l}{ApolloScape (3s)} & \multicolumn{2}{l}{NGSIM (5s)} \\ \cline{2-5} 
                        & ADE               & FDE               & ADE            & FDE            \\ \hline
Enc-Dec(LSTM)           & 2.24              & 8.25              & 6.86           & 10.02          \\
CS-LSTM                 & 2.14             & 11.69             & 7.25           & 10.05         \\
TraPHic                 & 1.28              & 11.67             & 5.63           & 9.91        \\
Social-GAN              & 3.98              & 6.75              & 5.65          & 10.29          \\
GRIP                    & 1.25              & 2.34              & 1.61           & 3.16           \\
Spectral                & 1.12              & 2.05              & 0.40           & 1.08           \\
RAFLTP (Ours)                    & \textbf{1.11}     & \textbf{2.01}     & \textbf{0.98}           & \textbf{2.04}           \\
DGInet                  & 0.99              & 1.74              & 0.37           & 1.01           \\ \hline
\end{tabular}
\label{perform}
\end{table}

\subsection{Visualization of Prediction Results}
In Fig.~\ref{viusal}, we use the NGSIM datasets to show several prediction results under different traffic conditions, where the circled vehicles are those that GRIP++ tries to predict, the purple solid line is the observed history, the red solid line is the future ground truth, the blue dashed line is the prediction result of our model (5 seconds), the green dashed line is the GRIP++ prediction (5 seconds). The range from -90 to 90 feet is the observation range. After observing the 3-second historical trajectories, our model predicts the trajectories over the 5-second horizon in the future. As can be seen in Fig.~\ref{viusal}, in various scenarios, when feeding the model with identical historical trajectories, our proposed model (blue dashed line) shows a closer match with the actual trajectories represented by the red solid line compared to GRIP++ (green dashed line), especially at the endpoints of the dashed lines. This further substantiates the superior predictive performance of our approach over GRIP++.

\subsection{Computation time}
The computational efficiency of the algorithm is a critical performance metric for autonomous vehicles. The computation time of our proposed approach is reported in Table~\ref{time}, implemented using PyTorch. We evaluate the prediction time for 1000 vehicles with batch sizes of 128 and 1, respectively. Our trajectory prediction model is an enhancement based on GRIP++. Although our computational time exhibits a delay compared to the GRIP++ model, it remains acceptable compared to other methods. Despite the latency incurred by our approach, it provides data privacy protection and higher prediction accuracy in return. Notably, our proposed PPO method for grouping vehicles operates on the server. The pre-trained PPO grouping model is deployed on the server, requiring only periodic experience replay\cite{liang2021ptrppo} and asynchronous updates, without impacting the prediction efficiency of the vehicles themselves.

\begin{table}[!t]
\centering
\caption{Computation time.}
\begin{tabular}{cccc}
\hline
\multirow{2}{*}{Scheme} &
  \multirow{2}{*}{Predicted \#} &
  \multirow{2}{*}{\begin{tabular}[c]{@{}c@{}}Times (s) \\ batch size 128\end{tabular}} &
  \multirow{2}{*}{\begin{tabular}[c]{@{}c@{}}Times (s) \\ batch size 1\end{tabular}} \\
         &      &      &        \\ \hline
CS-LSTM  & 1000 & 0.29 & 35.13  \\
GRIP     & 1000 & 0.05 & 6.33   \\
GRIP++   & 1000 & 0.02 & 1.62   \\
Spectral & 1000 & 1.10 & 120.20 \\
DGInet   & 1000 & 0.07 & 6.42   \\ \hline
RAFLTP (Ours)    & 1000 & 0.38 & 5.278  \\ \hline
\end{tabular}

\label{time}
\end{table}

\section{CONCLUSION}
This paper presents an innovative framework to address the distributed trajectory prediction problem. Within this framework, we leverage graph neural network tools and propose an asynchronous federated learning-based data sharing approach with an interpretable reputation quantization mechanism. Data providers share data structures under differential privacy constraints to ensure security. We implement PPO to categorize vehicles based on reputation levels, thereby optimizing the efficiency of FL aggregation. Experimental results demonstrate that the proposed data sharing scheme is more tolerant of bad nodes, enhances the security of the distributed trajectory prediction task, and significantly improves prediction accuracy. In future research, we plan to focus on integration with 6G networks and further enhancing efficiency.



 
%

\bibliographystyle{IEEEtran}
\bibliography{TPFL}

\newpage

 




\vfill

\end{document}